\documentclass[journal]{IEEEtran}
\usepackage{amsmath,amsfonts}
\usepackage[ruled,linesnumbered]{algorithm2e}
\usepackage{array}
\pdfoutput=1 % arxiv

\ifCLASSOPTIONcompsoc
\usepackage[caption=false, font=normalsize, labelfont=sf, textfont=sf]{subfig}
\else
\usepackage[caption=false, font=footnotesize]{subfig}
\fi
\usepackage{textcomp}
\usepackage{stfloats}
\usepackage{url}
\usepackage{verbatim}
\usepackage{graphicx}
\usepackage{cite}
\usepackage{amssymb}
\usepackage{booktabs}
\usepackage{algpseudocode}
\usepackage{multirow}
\usepackage{tabularx}
\usepackage{enumerate}
\usepackage{bigstrut}
\usepackage{dsfont}
\usepackage{float}
\usepackage{makecell}
\usepackage{ragged2e}
\usepackage[hidelinks,colorlinks,allcolors=black,pdfstartview=Fit]{hyperref}
\usepackage[capitalize]{cleveref}
\crefname{section}{Sec.}{Secs.}
\Crefname{section}{Section}{Sections}
\Crefname{table}{Table}{Tables}
\crefname{table}{Tab.}{Tabs.}
\newcolumntype{Z}{>{\centering\let\newline\\\arraybackslash\hspace{0pt}}X}
\begin{document}
\title{EARL: An Elliptical Distribution aided Adaptive Rotation Label Assignment for Oriented Object Detection in Remote Sensing Images}
\author{Jian Guan, \IEEEmembership{Member, IEEE}, Mingjie Xie, Youtian Lin, Guangjun He, Pengming Feng, \IEEEmembership{Member, IEEE}
% <-this % stops a space
  \thanks{This work was partly supported by the Natural Science Foundation of Heilongjiang Province under Grant No. YQ2020F010. (Corresponding author: Pengming Feng.)}
  \thanks{Jian Guan and Mingjie Xie are with the Group of Intelligent Signal Processing (GISP), College of Computer Science and Technology, Harbin Engineering University, Harbin 150001, China (e-mail: j.guan@hrbeu.edu.cn; xiemingjie@hrbeu.edu.cn).}
  \thanks{Youtian Lin is with the School of Electronics and Information Engineering, Harbin Institute of Technology, Harbin 150001, China (e-mail: linyoutian.loyot@gmail.com).}
  \thanks{Guangjun He and Pengming Feng are with the State Key Laboratory of Space-Ground Integrated Information Technology, CAST, Beijing 100095, China (e-mail: hgjun\_2006@163.com; p.feng.cn@outlook.com).}
}
\maketitle
\begin{abstract}
Label assignment is a crucial process in object detection, which significantly influences the detection performance by determining positive or negative samples during training process. However, existing label assignment strategies barely consider the characteristics of targets in remote sensing images (RSIs) thoroughly, e.g., large variations in scales and aspect ratios, leading to insufficient and imbalanced sampling and introducing more low-quality samples, thereby limiting detection performance. To solve the above problems, an Elliptical Distribution aided Adaptive Rotation Label Assignment (EARL) is proposed to select high-quality positive samples adaptively in anchor-free detectors. Specifically, an adaptive scale sampling (ADS) strategy is presented to select samples adaptively among multi-level feature maps according to the scales of targets, which achieves sufficient sampling with more balanced scale-level sample distribution. In addition, a dynamic elliptical distribution aided sampling (DED) strategy is proposed to make the sample distribution more flexible to fit the shapes and orientations of targets, and filter out low-quality samples. Furthermore, a spatial distance weighting (SDW) module is introduced to integrate the adaptive distance weighting into loss function, which makes the detector more focused on the high-quality samples. Extensive experiments on several popular datasets demonstrate the effectiveness and superiority of our proposed EARL, where without bells and whistles, it can be easily applied to different detectors and achieve state-of-the-art performance.
The source code will be available at: \url{https://github.com/Justlovesmile/EARL}.

\end{abstract}
\section{Introduction}
\label{sec:1}
\IEEEPARstart{O}{bject} detection in remote sensing images (RSIs), which aims to determine the locations and categories of the interested targets, has become one of the most common and potential image interpretation steps for numerous applications\cite{deng2017toward,wang2019enhancing,shi2020orientation,yu2020orientation}, such as maritime rescuing\cite{varga2022seadronessee}, urban planning\cite{ma2018mobile} and traffic management\cite{asha2018vehicle}.

In the past few years, substantial methods\cite{ren2015faster,redmon2016you,retinanet,tian2019fcos,duan2019centernet} have made significant progress for generic object detection that are designed for nature images, which have greatly promoted the development of object detection in RSIs\cite{wang2020learning,dai2022ace,cheng2022anchor,cheng2021prototype} and attract wide attention from researchers.
However, object detection in RSIs is still facing challenges from the characteristics of targets, such as arbitrary orientations and large variations in scales and aspect ratios.
To tackle these challenges, current detectors for RSIs \cite{lin2019ienet,qin2021mrdet,yu2022object, han2021align,yang2019r3det} often enhance orientation prediction by incorporating additional branches and alleviate scale variations by adopting multi-level feature maps\cite{lin2017feature,zhu2019fsaf,hou2022refined,wang2020frpnet}.
However, it usually faces the crucial issue of selecting training (i.e., positive) samples at reasonable pixel-wise locations (i.e., spatial assignment) on appropriate levels of feature maps (i.e., scale assignment), namely label assignment \cite{zhu2020autoassign}.

As a fundamental and crucial process in object detection, label assignment significantly influences the detection performance\cite{zhang2020bridging}. However, existing detectors for RSIs often adopt label assignment strategies migrated from generic detectors, such as maximum intersection over union (IoU) based strategies  \cite{qin2021mrdet,yang2019r3det,yang2019scrdet} and center-based strategies \cite{cheng2022anchor,shi2021canet,feng2020toso}, which are often unsuitable for object detection in RSIs due to the large variations in scales and aspect ratios of targets, resulting in insufficient and imbalanced sampling and introducing more low-quality samples (e.g., samples located on the background or near the boundaries), thereby limiting detection performance.

\begin{figure}[!tbp]
    \centering
    \includegraphics[width=1.0\linewidth]{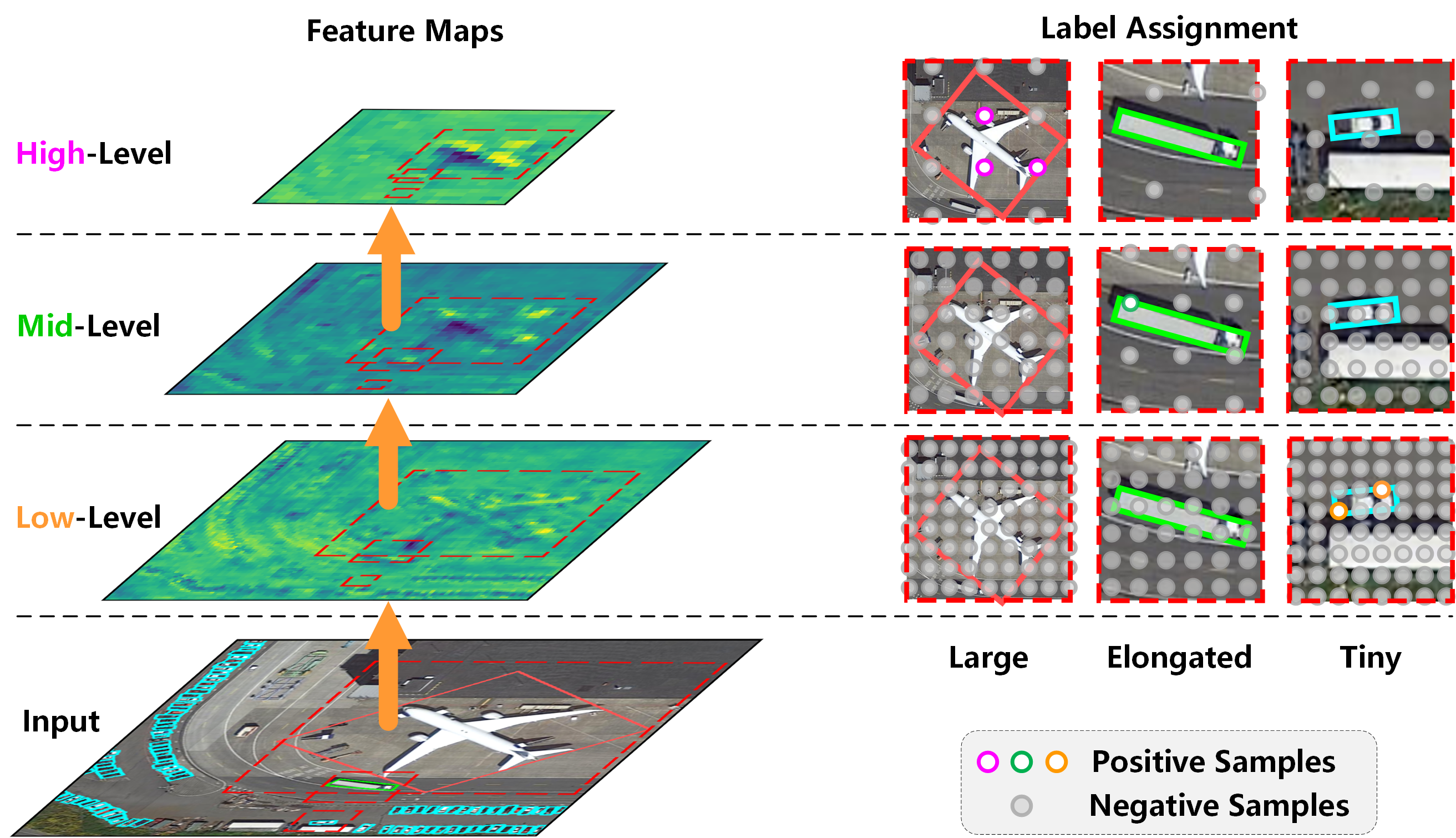}
     \caption{Illustration of insufficient sampling by using fixed scale assignment strategy, where targets with extreme scales and large aspect ratios, i.e., tiny car, elongated bus and large airplane, are taken for example. Each point in red dashed boxes denotes the location of pixel-wise sample on the input image. Here, only high-level and low-level samples are assigned to large and tiny targets, respectively, thereby resulting in only few or even no positive samples assigned to the targets with extreme scales and large aspect ratios. Best viewed in colour.}
     \label{fig:Imbalance}
\vspace{-0.2cm}
\end{figure}

Concretely, existing methods (e.g., AOPG \cite{cheng2022anchor} and GGHL \cite{huang2022general}) often employ the fixed scale assignment strategy, which heuristically assigns samples on a specific level of feature maps to the target decided by the predefined fixed threshold. However, this heuristic approach often results in insufficient and imbalance sampling, especially in RSIs. 
As illustrated in Fig.~\ref{fig:Imbalance}, few samples on the specific level of feature map are assigned to the targets with extreme scales and large aspect ratios (i.e., tiny car, elongated bus and large airplane), whereas for samples on  other levels of feature maps, despite being located on the target itself, they are not assigned to the target (i.e., treated as negative samples), due to the fixed threshold. In addition, this approach often aggravates the scale-level sampling imbalance, when the target scales are unbalanced, especially in RSIs, where small targets are far more numerous than large targets. Consequently, the selected positive samples are often clustered in the low-level of feature maps, thus overwhelming the sufficient attention to the targets with other scales and introducing the scale-level bias.
In contrast, other methods (e.g., FSDet\cite{yu2022object}) assign all samples across all levels of feature maps within the bounding box to the target, which often introduce large amounts of low-quality samples (e.g., background noise) and ambiguous samples (i.e., samples that are assigned to multiple targets simultaneously\cite{tian2019fcos}), especially for targets with overlapping relationships, such as harbours and ships, that would disturb the training process. Therefore, a more flexible method is required for sample selection on multi-level feature maps.

\begin{figure}[tbp]
    \centering
    \includegraphics[width=0.75\linewidth]{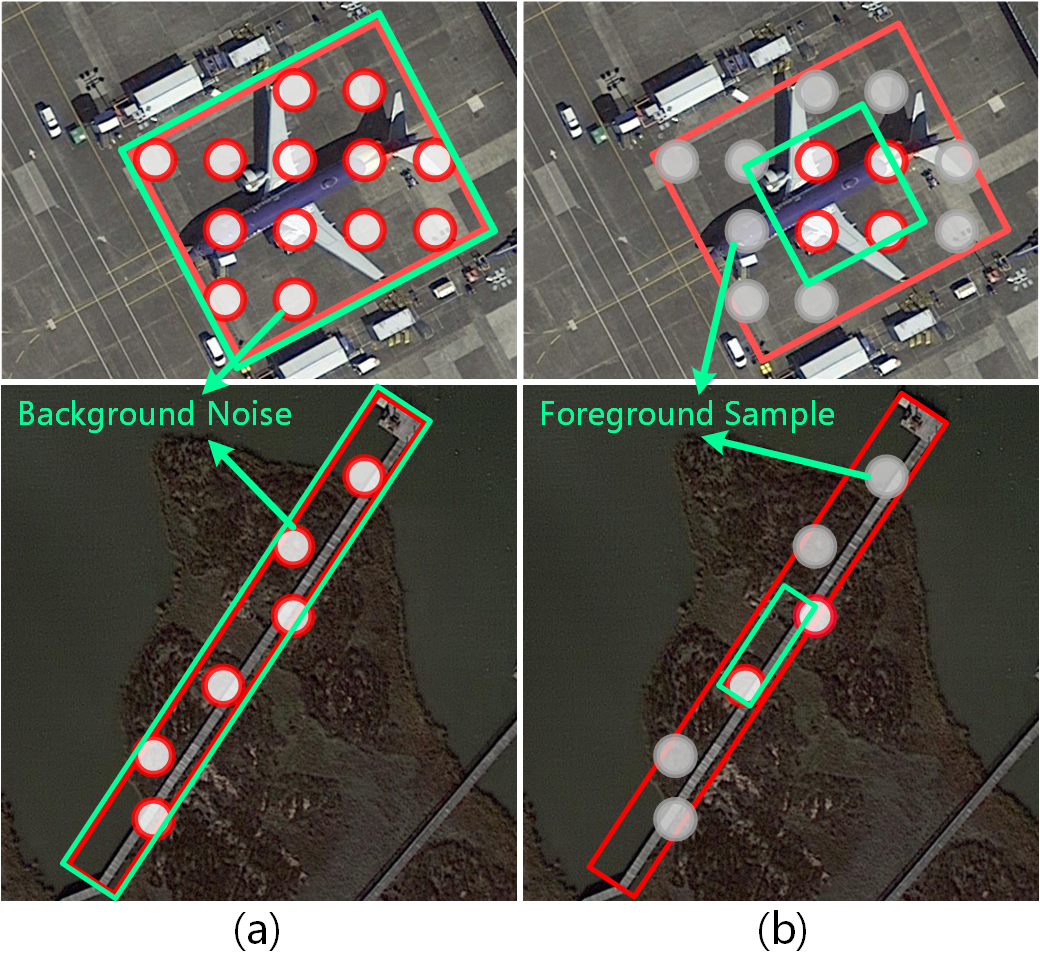}
    \caption{Illustration of the existing label assignments with different sampling ranges for spatial assignment. Here, (a) denotes using rectangle bounding box that introduces more background noise; (b) denotes using central area that discards more foreground samples.}
    \label{fig:backgroundissue}
\vspace{-0.2cm}
\end{figure}

In addition, existing detectors usually use rectangle bounding box\cite{feng2020toso,yu2022object} as the sampling range in spatial assignment, and tend to introduce more low-quality samples from the background of rectangle bounding box, as illustrated in Fig.~\ref{fig:backgroundissue}~(a). In contrast, center-based strategies \cite{zhang2020bridging,shi2021canet,cheng2022anchor} can improve the quality of selected samples by only sampling central area of targets. However, it still rarely takes into account the large variations in aspect ratios of targets in RSIs, which leads to many foreground samples being misclassified as negatives and results in insufficient sampling, as shown in Fig.~\ref{fig:backgroundissue}~(b). Hence, a more reliable spatial assignment strategy is necessary for detectors to capture positive samples with higher quality.

To address the above problems, this work proposed an elliptical distribution aided adaptive rotation label assignment (EARL) strategy to select high-quality positive samples adaptively in orientation anchor-free detectors and yield better performance. 
Specifically, an adaptive scale assignment (ADS) strategy is proposed to assign top-$k$ samples from high to low level of feature maps adaptively according to the scales of targets, i.e., assigning more high-level samples and fewer low-level samples to the large targets, and conducting the opposite for the small targets. Therefore, it can achieve adaptive and sufficient sampling for extreme targets with more balanced scale-level sample distribution. 
In addition,  a dynamic elliptical distribution aided sampling (DED) strategy is presented to select samples following an elliptical distribution, which can adjust sampling range dynamically according to the aspect ratios of targets, i.e., the sampling range tends to be a contracted circular distribution when the shape of target is close to a square, and an inner tangent ellipse when the shape is elongated, which can achieve more flexible sample distribution to fit the shapes and orientations of targets and filter out low-quality samples.
Moreover, a spatial distance weighting (SDW) module is introduced to assign different weights to samples at different pixel-wise locations, e.g., assigning larger weights to the samples close to the center of target, and smaller weights to the samples near the boundaries, which allows the detector to focus on the high-quality samples, thereby further enhancing the detection performance. 
Our proposed EARL can be easily deployed to different detectors to achieve better performance, where a simple anchor-free detector, R-FCOS \cite{tian2019fcos}, and an advanced detector, RTMDet-R \cite{lyu2022rtmdet}, with the minimal modification are employed to verify the effectiveness of our method. Extensive experiments on several popular RSIs datasets demonstrate the effectiveness and superiority of our proposed EARL.

In summary, the contributions of this work are summarized as follows:
\begin{enumerate}
\item An ADS strategy is proposed to select training samples adaptively among multi-level feature maps, which achieves sufficient sampling with more balanced scale-level sample distribution.
\item A DED strategy is proposed to make the sample distribution more flexible to fit the shapes and orientations of the targets while filtering out low-quality samples.
\item An SDW module is introduced to enhance the high-quality sample selection by integrating the adaptive distance weighing into loss function, and which can further improve the detection performance.
\item Extensive experiments on three challenging datasets were conducted. The results verify the effectiveness of each component of the proposed EARL, and demonstrate that our strategy can be applied to different baseline detectors and improve the detection performance. 
\end{enumerate}

\section{Related Work}
\label{sec:2}

\subsection{Oriented Object Detection in RSIs}
Due to its wide range of application scenarios, oriented object detection in RSIs has developed rapidly. Similar to generic detectors, recent methods can be divided into two categories, i.e., anchor-based detectors and anchor-free detectors. To detect targets with arbitrary orientations, some anchor-based detectors\cite{liu2017rotated,zhang2018toward,yang2018automatic}, which are built on the two-stage framework, i.e., Faster R-CNN\cite{ren2015faster}, densely preset multiple anchors with different scales, aspect ratios and angles for better regression while introducing heavy anchor-related computations. After that, RoI-Transformer\cite{ding2019learning} proposed an efficient region of interests (RoI) learner to transform horizontal proposals to rotated ones. Gliding Vertex\cite{xu2020gliding} described oriented targets by horizontal bounding boxes (HBB), which further avoided numerous anchors with multiple angles. Then, DODet\cite{cheng2022dual} was proposed to deal with the spatial misalignment  between horizontal proposals and oriented targets, as well as the feature misalignment between classification and localization.
Whereas R$^3$Det\cite{yang2019r3det} and S$^2$A-Net\cite{han2021align} followed the one-stage schema to balance accuracy and efficiency, and alleviated the feature misalignment issue. However, these methods still take a long time during training and inference\cite{yu2022object}.

To maintain high efficiency, anchor-free detectors are designed to avoid the usage of anchors. For instance, TOSO\cite{feng2020toso} directly regressed the surrounding HBB and transformation parameters to present rotated targets. IENet\cite{lin2019ienet} was built upon the classic one-stage anchor-free framework, i.e., FCOS, adding an additional branch with the interactive embranchment module to regress angles. Later, BBAVector\cite{yi2021oriented}, O$^2$-DNet\cite{wei2020oriented} and ACE\cite{dai2022ace} employed a simplified architecture and designed new oriented bounding box (OBB) representation methods.
However, these anchor-free detectors still not perform the state-of-the-art results compared with anchor-based detectors, and most of them are more likely to improve the architecture while not considering the training strategy, especially the label assignment strategy, which has been proven to be the essential difference between these two kinds of detectors\cite{zhang2020bridging}.

\begin{figure*}[htbp]
    \centering
    \includegraphics[width=0.9\linewidth]{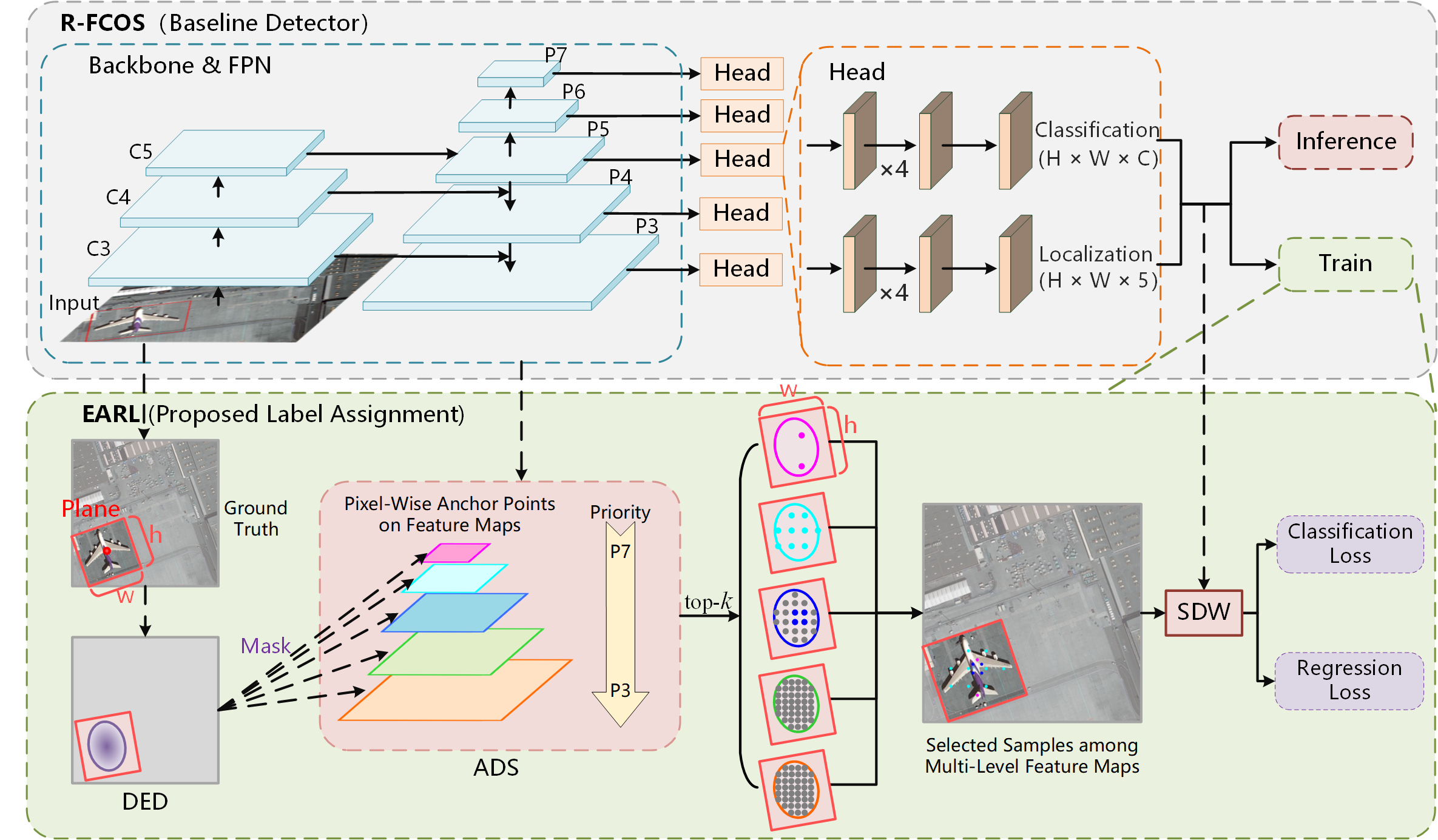}
    \caption{Overview of the proposed EARL with the deployment on a simple baseline detector. The upper gray block shows the baseline architecture, where R-FCOS is taken for example. The lower green block presents our proposed EARL strategy. The DED strategy is employed based on the shape of the target and generates foreground masks on each level of feature maps outlining area to select candidate samples. The candidate samples will be sorted from the highest to lowest level of feature maps where the samples in the same level will sorted by the distance from the center point of ground truth bounding box. Only the first top-$k$ samples are selected as positive samples (coloured points). Best viewed in colour.}
    \label{fig:Architecture}
\end{figure*}

\subsection{Label Assignment Strategies}
Label assignment is a crucial component for convolutional neural network (CNN) based detectors to learn feature distribution of targets by determining positive or negative samples. Meanwhile, as the current detectors commonly adopt multi-scale feature maps to alleviate the scale variations, label assignment needs to simultaneously assign labels to samples at different spatial locations (spatial assignment) on different levels of feature maps (scale assignment)\cite{zhu2020autoassign}. Most anchor-based detectors\cite{ren2015faster,retinanet,qin2021mrdet} assign the labels to anchors among multi-level feature maps by comparing the preset thresholds and IoU between anchors and ground truth bounding boxes. However, this strategy often involves many hyperparameters to adjust the scales and aspect ratios of anchors based on the datasets\cite{zhang2021learning}. Whereas anchor-free methods\cite{tian2019fcos,kong2020foveabox,xiao2020axis} often select samples inside the ground truth bounding box or the central area of targets for spatial assignment, and choose samples on a specific level of feature maps according to the predefined target size threshold for scale assignment. However, above strategies, which depend on the heuristic rules, may not be optimal enough. 

Thus, recent works provide dynamic label assignment strategies, which allow models to learn to select samples by itself. For instance, FSAF\cite{zhu2019fsaf} dynamically assigned targets to the suitable feature levels based on computed loss. ATSS\cite{zhang2020bridging} proposed an adaptive training sample selection method by adjusting the IoU threshold according to the statistics of targets, which also proposed the center sampling strategy to improve the quality of positive samples. Autoassign\cite{zhu2020autoassign} used two weighting modules to adjust the category-specific prior distribution according to the appearances of targets. As for oriented object detection, SASM\cite{hou2022shape} was motivated by the idea of ATSS and designed a dynamic IoU threshold that incorporates shape information. RTMDet\cite{lyu2022rtmdet} proposed a dynamic soft label assignment based on SimOTA\cite{ge2021yolox}. GGHL\cite{huang2022general} used a Gaussian heatmaps to define positive samples according to the shape and direction properties of targets. AOPG\cite{cheng2022anchor} applied a region assignment scheme to select positive samples within the shrinked OBB area. Oriented RepPoints\cite{li2022oriented} used an adaptive quality assessment and sample assignment scheme. FSDet\cite{yu2022object} assigned a soft weight to each sample among all levels of feature maps. However, the label assignments used by these methods are rarely consider the specific characteristics of targets in RSIs, i.e., large variations in scales and aspect ratios, resulting in insufficient and imbalanced sampling and often introducing large amounts of low-quality samples. Therefore, this study focuses on designing an adaptive label assignment strategy for orientation anchor-free detectors by taking into account the important characteristics in RSIs to address aforementioned problems.

\section{The Proposed EARL Method}
\label{sec:3}
In this section, the architecture of our simple baseline anchor-free detector is firstly presented. Then, the three components of the proposed EARL, i.e., ADS strategy, DED strategy and SDW module, are introduced in detail.
\subsection{Architecture of the Simple Baseline Anchor-free Detector}
\label{sec:3-1}
To demonstrate the effectiveness of the proposed label assignment strategy, in this paper, a simple but classical anchor-free architecture, namely R-FCOS, is employed as our baseline detector. Compared with FCOS\cite{tian2019fcos}, R-FCOS removes centerness branch to be more compact and achieves more simplicity and efficiency, which only contains classification branch and regression branch for oriented object detection. 

As shown in Fig.~\ref{fig:Architecture}, the upper gray block shows the baseline architecture, which consists of the backbone network, feature pyramid network (FPN) and prediction heads.  Let $\mathbf{C}_\ell$ and $\mathbf{P}_{\ell}\in \mathbb{R}^{H_\ell \times W_\ell \times \mathcal{C}_\ell}$ be the feature maps from backbone network and FPN, respectively, where $\ell$ is the layer index, $H_\ell \times W_\ell$ represents the size of feature maps and $\mathcal{C}_\ell$ denotes the number of feature channels.
In this work, following FCOS, five levels of multi-scale feature maps $\{\mathbf{P}_3,\mathbf{P}_4,\mathbf{P}_5,\mathbf{P}_6,\mathbf{P}_7\}$ are used, where $\mathbf{P}_3,\mathbf{P}_4$ and $\mathbf{P}_5$ are generated from $\mathbf{C}_3,\mathbf{C}_4$ and $\mathbf{C}_5$, respectively. Whereas $\mathbf{P}_6$ and $\mathbf{P}_7$ are generated via up-sampling $\mathbf{P}_5$ and $\mathbf{P}_6$, respectively.

\begin{figure}[tbp]
    \centering
    \includegraphics[width=0.5\linewidth]{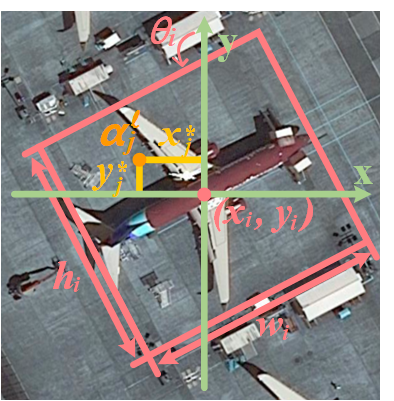}
    \caption{Illustration of ground truth. $(x_i,y_i,w_i,h_i,\theta_i)$ are the center, width, height and angle of $g_i$, and $(x_j^*,y_j^*)$ denotes the offset between $\alpha_j^\ell$ and the center of $g_i$. Best viewed in colour.}
    \label{fig:groundtruth}
\end{figure}

After obtaining the feature maps from FPN, the prediction heads with two fully convolutional subnets are employed to predict the categories and regressions for each location on feature maps.
Specifically, given a set of ground truths $\mathcal{G}=\{g_{i}\}_{i=1}^n$ in the input image, where $n$ denotes the total number of ground truths and $i$ is the index, each $g_i$ in $\mathcal{G}$ is represented by $g_i=(x_i,y_i,w_i,h_i,\theta_i,c_i)$, as shown in Fig.
~\ref{fig:groundtruth}, where $(x_i, y_i)$ is the coordinate of center point of $g_i$, $w_i$ and $h_i$ represent the width and height of $g_i$, and $\theta_i\in[0^\circ,180^\circ)$ represents the counterclockwise angle between the y-axis and long side, and $c_i$ is the class label that the ground truth belongs to.

To generate prediction, we adopt $\mathcal{A}_\ell=\{\alpha^{\ell}_{j}\}_{j=1}^{H_\ell \times W_\ell}$ to denote the set of anchor points, which are the pixel-wise locations on $\mathbf{P}_{\ell}$. For each anchor point $\alpha^{\ell}_{j}$, it can be mapped back to the input image via:
\begin{equation}
  \begin{split}
    x_j = \lfloor \frac{s_\ell}{2} \rfloor+\tilde{x}^{\ell}_j{s_\ell}\\
    y_j = \lfloor \frac{s_\ell}{2} \rfloor+\tilde{y}^{\ell}_j{s_\ell}
  \end{split}
\end{equation}
where $(\tilde{x}^{\ell}_j,\tilde{y}^{\ell}_j)$ and $(x_j,y_j)$ represent the locations of $\alpha_j^\ell$ on $\mathbf{P}_\ell$ and the input image, respectively. $s_\ell$ denotes the stride on $\mathbf{P}_\ell$ and $\lfloor\cdot\rfloor$ denotes the round-down operator. Here, we directly regress the target bounding box at each $\alpha_j^{\ell}$. Note that, the label assignment strategies are often employed to decide whether each $\alpha_j^{\ell}$ is a positive or negative sample, and allow the network to select higher-quality samples for training.

Then, if $\alpha_j^{\ell}$ is a positive sample and assigned to the ground truth, e.g., $g_i$, its class label $c^*_j$ equals to $c_i$ and it has a 5-dimensional vector $t_j^*=(x^*_j,y^*_j,w^*_j,h^*_j,\theta^*_j)$ being the regression targets for the location, where $x^*_j$ and $y^*_j$ are the coordinate offsets between the center point of $g_i$ and $\alpha_j^{\ell}$, as shown in Fig.~\ref{fig:groundtruth}, which can be formulated as:
\begin{equation}
  \begin{split}
    x^*_j = x_i - x_j\\
    y^*_j = y_i - y_j
  \end{split}
\end{equation}
$w^*_j$, $h^*_j$ and $\theta^*_j$ equal to the width, height and angle of $g_i$, respectively.

During inference, the prediction heads will directly generate classification and regression for each anchor point and output the classification scores and locations after post-processing, hence our proposed strategy is inference cost-free.

The above represents the pipeline of the simple baseline detector, as illustrated in Fig. \ref{fig:Architecture}. This simple baseline detector is only used for illustration to show how our proposed EARL strategy works on the baseline detector for sample selection by simply replacing its original label assignment strategy.

Note that, the proposed EARL strategy can be also easily deployed on other baseline detector in the same way. Specifically, we also equipped it with an advanced anchor-free detector, i.e., RTMDet-R \cite{lyu2022rtmdet}, for performance evaluation to demonstrate the effectiveness of our study in the experiments.

In the next subsections, the components of our proposed EARL, which are given in the lower green block of Fig. \ref{fig:Architecture}, will be introduced in detail.

\subsection{Adaptive Scale Sampling Strategy}
\label{sec:3-2}
Different from the existing scale assignment strategies, the simple but novel ADS strategy is proposed to select positive samples adaptively according to the scales of targets. Its implementation can be summarized as sampling sequentially from high to low-level feature map, e.g., from  $\mathbf{P}_7$ to $\mathbf{P}_3$, as illustrated in Fig. \ref{fig:Architecture}.

Specifically, for each $g_i$ in the input image, a determined function $\tau(\cdot)$ is used to indicate whether each $\alpha^{\ell}_j$ is inside a certain region of $g_i$ when mapped back to the input image, which can be simply understood as selecting samples inside the mask region on each $\mathbf{P}_\ell$, as shown in Fig.~\ref{fig:Architecture}.
Then, if $\alpha^{\ell}_j$ is judged to be inside the mask region by $\tau(\cdot)$, it will be selected into the candidate samples set $\mathcal{R}_\ell$, where $\tau(\cdot)$ will be discussed in more detail in Section \ref{sec:3-3}.

\begin{algorithm}[t]
  \SetKwInOut{Input}{Input}\SetKwInOut{Output}{Output}
  \caption{Algorithm of ADS strategy}\label{alg:cap}
  \Input{The set of ground truths $\mathcal{G}$ in the input image, the set of all anchor points $\mathcal{A} = \{\mathcal{A}_\ell\}_{\ell=3}^{\ell=7}$ among all feature levels and the sample number $k$}
  \Output{The set of positive samples $\mathcal{P}$ and negative samples $\mathcal{N}$}
  \BlankLine
  \For{each ground truth $g_i \in \mathcal{G}$}{
    \emph{Build an empty set for positive samples $\mathcal{P}\gets\varnothing$}\;
    \For{layer index $\ell\gets7$ to $3$}{
      \If{$k > 0 $}{
        \emph{Build an empty set for candidate samples $\mathcal{R}_{\ell}\gets\varnothing$}\;
        \For{each anchor point $\alpha_j^\ell \in \mathcal{A}_{\ell}$}{
            \If{$\tau(\alpha_j^\ell, g_i)$ is True}{
                \emph{$\mathcal{R}_{\ell} = \mathcal{R}_{\ell} \cup \alpha_j^\ell$}
            }
        }
        \emph{$\tilde{n}_{\mathcal{S}}^\ell=min(k,\tilde{n}_{\mathcal{R}}^\ell)$}\;
        \emph{$\mathcal{S}_{\ell}\gets$ select top-$\tilde{n}_{\mathcal{S}}^\ell$ anchor points from $\mathcal{R}_{\ell}$}\;
        \If{$\tilde{n}_{\mathcal{S}}^\ell > 0$}{
            \emph{$k = k - \tilde{n}_{\mathcal{S}}^\ell$}\;
            \emph{$\mathcal{P}=\mathcal{P} \cup \mathcal{S}_{\ell}$}\;
        }
      }
    }
  }
  \emph{$\mathcal{N}=\mathcal{A}-\mathcal{P}$}\;
  \emph{\Return {$\mathcal{P}$, $\mathcal{N}$}}\;
\end{algorithm}

When the candidate sample selection is applied from $\mathbf{P}_7$ to $\mathbf{P}_3$, we only keep the $k$ samples that are closest to the center point of $g_i$ as positive samples based on L2 distance, which is denoted as top-$k$ in this paper. Here $k$ is a hyperparameter that ensures a balanced number of positive samples for each target as possible, which is suggested in \cite{retinanet} that unbalanced positive samples will decrease the performance, and L2 distance indicates the Euclidean Distance.
In details, let $\mathcal{S}_\ell$ denote selection set on $\mathbf{P}_\ell$ and $\tilde{n}_{\mathcal{S}}^\ell$ denote the number of selected samples on $\mathbf{P}_{\ell}$, which is formulated as:
\begin{equation}
  \begin{split}
    \tilde{n}_{\mathcal{S}}^\ell = min(k,\tilde{n}_{\mathcal{R}}^\ell)
  \end{split}
\end{equation}
where $\tilde{n}_{\mathcal{R}}^\ell$ denotes the number of candidate samples on $\mathbf{P}_{\ell}$. Finally, we subtract $\tilde{n}_{\mathcal{S}}^\ell$ from $k$ and add the selected candidate samples to the positive samples set $\mathcal{P}$, after which we repeat this operation if $k$ is greater than 0. Algorithm~\ref{alg:cap} specifies the proposed method for positive sample selection.

\begin{figure}[t]
    \centering
     \includegraphics[width=1.0\linewidth]{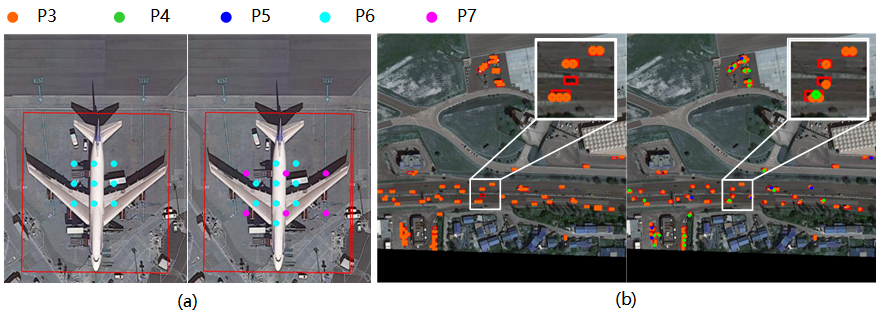}
     \caption{Illustration of FSA strategy (left column) and ADS strategy (right column) for sampling positive samples of targets with different scales. Best viewed in colour.}
     \label{fig:ADS}
\end{figure}

As a result,  more high-level samples and fewer low-level samples are assigned to the large targets, and the opposite is conducted for the small targets, according to the scales of targets, while the number of ambiguous samples can be reduced adaptively. In addition, if an anchor point is still assigned to multiple ground truths, it is forced to be assigned to the one with the longest side. Besides, as shown in Fig.~\ref{fig:ADS}, if a ground truth escapes from any samples, we assign it with the closest unassigned anchor point, which ensures the information from each target can be used reasonably and partially alleviates the insufficient sampling problem.

\begin{figure}[t]
    \centering
     \includegraphics[width=1.0\linewidth]{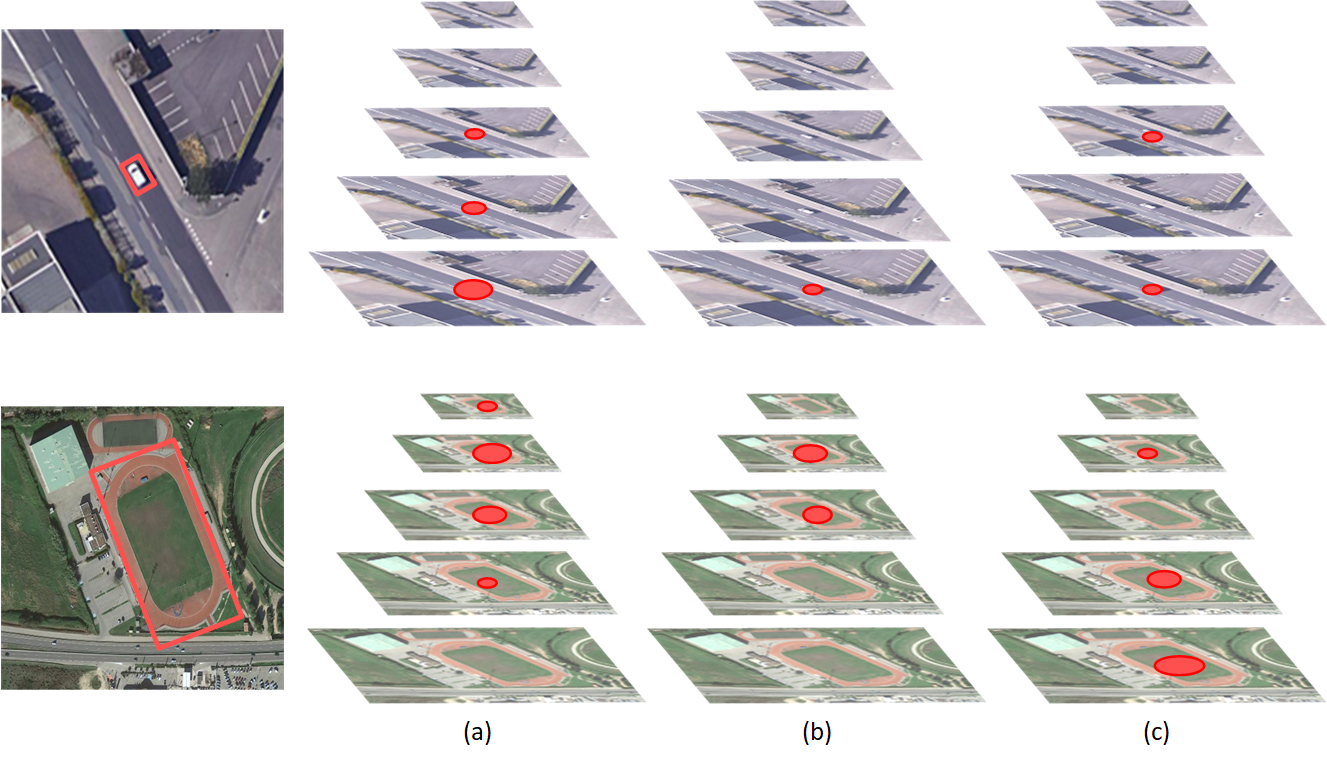}
     \caption{Illustration of the comparison about the number of positive samples by using different sampling strategies on multi-level feature maps for targets with different scales. The size of the red range on the multi-level feature maps represents the number of selected positive samples. (a) ADS strategy. (b) FSA strategy. (c) TKT strategy.}
     \label{fig:AdaptiveSampleAssignment}
\end{figure}

In this way, the sequentially adaptive sampling operation of ADS can obtain a better distribution of positive samples when compared with the fixed scale assignment (FSA) strategy and the top-$k$ samples for each target (TKT) strategy (i.e., ADS without the sequential selection). As shown in Fig.~\ref{fig:AdaptiveSampleAssignment}, FSA strategy constrains the level distribution of positive samples. Whereas TKT strategy selects samples clustered at the bottom levels of feature maps, resulting in the imbalanced scale-level sample distribution. Therefore, above strategies will lead to the scale-level bias, especially when the target scale distribution is unbalanced, which is a common phenomenon in RSIs. In contrast, our proposed ADS can alleviate the problems mentioned above by sequentially sampling from high to low level of feature maps adaptively according to the scale variation of targets in RSIs.

\subsection{Dynamic Elliptical Distribution aided Sampling Strategy}
\label{sec:3-3}
\begin{figure}[t]
    \centering
     \includegraphics[width=1.0\linewidth]{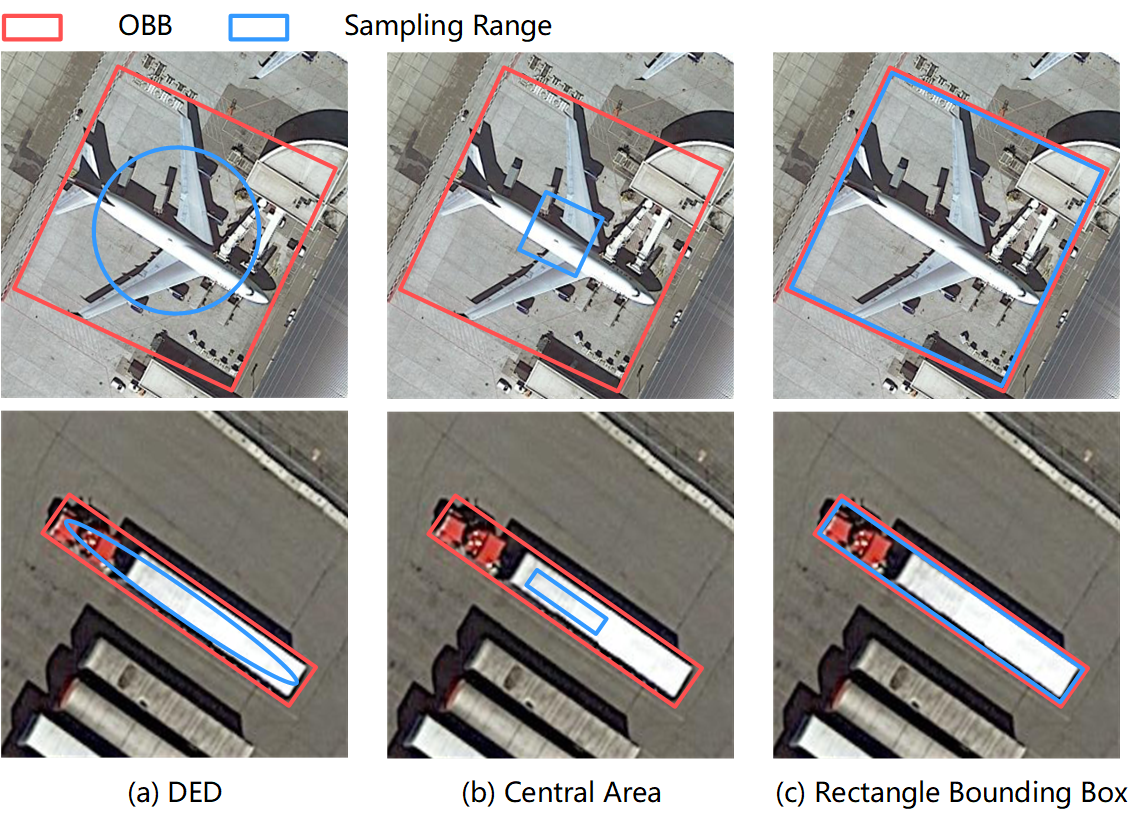}
     \caption{Illustration of the comparison of different sampling range in spatial assignment. Best viewed in colour.}
     \label{fig:ellipseSampling}
\end{figure}
In the Section \ref{sec:3-2}, the Algorithm~\ref{alg:cap} has a determined function $\tau(\cdot)$ that takes the ground truth $g_i$ and the anchor point $\alpha_j^\ell$ as input, and returns the true or false value depending on whether $\alpha_j^\ell$ is inside the certain region of $g_i$ or not. Here, the determined function achieves spatial assignment by constraining the sampling range. Concretely, existing methods \cite{yu2022object,feng2020toso} often adopt the central area  or the rectangle bounding box as the sampling range for spatial assignment, as shown in Fig.~\ref{fig:ellipseSampling} (b) and (c), respectively. 
However, these methods rarely consider the characteristics of targets in RSIs, which either introduce more background noise or result in insufficient sampling.

To this end, the proposed DED strategy selects samples following an elliptical distribution as illustrated in Fig. \ref{fig:ellipseSampling} (a), which can adjust sampling range dynamically according to the aspect ratios of targets, i.e., the sampling range tends to be a contracted circular distribution when the shape of target is close to a square, and an inner tangent ellipse when the shape is elongated. Such that it can achieve more flexible sample distribution to fit the shapes and orientations of targets.

Suppose that given a ground truth $g_i$ and an anchor point $\alpha_j^\ell$, which is mapped back to the input image, the function $\tau(\cdot)$ can be formulated  as:
\begin{equation}
    \tau(\cdot) = \begin{cases}
        \text{true} ,& \frac{a^2}{(\frac{1}{2}w_i)^2} + \frac{b^2}{(\frac{1}{2}h_i)^2} < \xi \\
        \text{false} ,& \text{otherwise}
    \end{cases}
    \label{eq:ded}
\end{equation}
where $a$ and $b$ are calculated by the angle of $g_i$ and the offsets of coordinates between $g_i$ and $\alpha_j^\ell$, which are formulated respectively as:
\begin{equation}
  \begin{split}
    a =  x^*_j\cos{\theta_i}+y^*_j\sin{\theta_i}\\
    b =  x^*_j\sin{\theta_i}-y^*_j\cos{\theta_i}
  \end{split}
\end{equation}
the ratio factor $\xi$ is an adaptive threshold determined by the target shape that controls the range of elliptical distribution, and to select high-quality samples with less background noise. Here, $\xi$ is calculated as follows:
\begin{equation}
    \xi = 1 - \frac{min(h_i,w_i)}{2\times max(h_i,w_i)}
    \label{eq:xi}
\end{equation}

In this way, by considering the shapes and orientations of the targets, our proposed DED strategy can avoid introducing too much background noise, which is more suitable for object detection in RSIs.

\subsection{Spatial Distance Weighting Module}
\label{sec:3-4}
To further mitigate the effect of low-quality predictions generated from the anchor points far away from the center of ground truth, instead of employing extra prediction branch such as centerness branch\cite{tian2019fcos} or IoU prediction branch\cite{yu2022object}, SDW module is proposed to enhance the high-quality sample selection by assigning different weights to the samples at different pixel-wise locations.

For each positive sample $\alpha_j^\ell\in\mathcal{P}$, SDW module calculates weight $\mathcal{W}_j$ according to the scale of $g_i$ that it belongs to, and the L2 distance between $\alpha_j^\ell$ and center point of $g_i$, which is represented as $d_{\alpha_j^\ell,g_i}$. For classification, $\mathcal{W}_j$ is used to weight the probability for the corresponding category. For box regression, we first establish the regression loss and then multiply by $\mathcal{W}_j$. Here, $\mathcal{W}_j$ is formulated as:
\begin{equation}
\label{eq:7}
\mathcal{W}_j = \frac{1-\frac{d_{\alpha_j^\ell,g_i}}{m}}{max(1-\frac{d_{\mathcal{P},g_i}}{m})}
\end{equation}
where $m=max(\frac{h_i}{2},\frac{w_i}{2})$, and $d_{\mathcal{P},g_i}$ is the set of all the distance of positive samples for $g_i$. The weight is normalized as in Eq. (\ref{eq:7}) to make sure that the highest weight is given to the closest positive sample, and achieves adaptive distance weighting, i.e., assigning larger weights to the samples close to the center of the target, and smaller weights to the samples near the boundaries.

Finally, $\mathcal{W}_j$ is applied to weight the classification probability and box regression loss. The total loss $\mathcal{L}_{total}$ is constructed with two types of loss, i.e., the classification loss $\mathcal{L}_{cls}$ and box regression loss $\mathcal{L}_{reg}$, denoted as
\begin{equation}
\begin{split}
    \mathcal{L}_{total} = \frac{1}{N_{pos}}\sum_\ell\sum_j\mathcal{L}_{cls}(\hat{c_j}, c^*_j, \mathcal{W}_j) \\
    + \frac{\lambda}{N_{pos}}\sum_\ell\sum_j\mathds{1}_{\{\alpha_j^\ell\in\mathcal{P}\}}\mathcal{L}_{reg}(\hat{t_j}, t^*_j, \mathcal{W}_j)
\end{split}
\end{equation}
where $\lambda$ is the penalty parameter to balance these two types of loss, $N_{pos}$ denotes the number of positive samples, which is used for normalization. $\mathds{1}_{\{condition\}}$ is the indicator function, which equals to 1 if the condition is satisfied, otherwise  0. The classification loss is formulated as:
\begin{equation}
    \label{eq:classification}
      \mathcal{L}_{cls}(\hat{c_j}, c^*_j, \mathcal{W}_j) =
      \begin{cases}
        - \delta(\mathcal{W}_j-\hat{c_j})^{\gamma}log(\hat{c_j}) ,& {c^*_j=1} \\
        - \delta(\hat{c_j})^{\gamma}log(1-\hat{c_j}) ,& \mbox{otherwise}
      \end{cases}
    \end{equation}
where $c^*_j\in\{0,1\}$ is the ground truth for classification and $\hat{c_j}\in[0,1]$ is the classification score from the network, $\delta$ and $\gamma$ are hyperparameters of the focal loss\cite{retinanet}. The box regression loss is formulated as:
\begin{equation}
    \label{eq:13}
    \begin{aligned}
    \mathcal{L}_{reg}(\hat{t_j}, t^*_j, \mathcal{W}_j)&= \mathcal{W}_j \times Smooth_{L1}(||\hat{t_j} - t^*_j||)
    \end{aligned}
\end{equation}
where $Smooth_{L1}(\cdot)$ is smooth L1 loss function as defined in \cite{girshick2015fast}.

With SDW module to enhance the high-quality sample selection, our proposed EARL strategy can further improve the detection performance.

\section{Experiments and Results}
\label{sec:4}
In this section, experiments on publicly available and challenging RSIs datasets are conducted to verify the effectiveness of our proposed EARL. Details of the experiment setup, parameter selection, ablation study, comparison with existing label assignment strategies and state-of-the-art methods are presented in the following subsections.
\subsection{Experiment Setup}
\subsubsection{Datasets}
DOTA\cite{xia2018dota} is a large-scale dataset for object detection in RSIs with OBB annotations for oriented targets, which contains 2,806 aerial images collected from different sensors and platforms. The fully annotated images contain 188,282 instances with a wide variety of scales, orientations and shapes, involving 15 common target categories: plane (PL), baseball diamond (BD), bridge (BR), ground track field (GTF), small vehicle (SV), large vehicle (LV), ship (SH), tennis court (TC), basketball court (BC), storage tank (ST), soccer ball field (SBF), roundabout (RA), harbor (HA), swimming pool (SP) and helicopter (HC).
In our experiments, DOTA is used following \cite{lin2019ienet}, where both training and validation sets are used for training, and the test set is employed for evaluating by submitting results to the evaluation server from the dataset provider. Due to the large size of images, following \cite{yang2019r3det}, all images from the datasets are cropped to $600\times600$ pixels with a stride of $450$ pixels for memory efficiency, and resized to $800\times800$ pixels during both training and inference.

DIOR-R\cite{cheng2022anchor} is a large-scale oriented object detection dataset containing 23,463 images and 192,518 instances. The dataset includes 20 common target categories: airplane (APL), airport (APO), baseball field (BF), BC, BR, chimney (CH), expressway service area (ESA), expressway toll station (ETS), dam (DAM), golf field (GF), GTF, HA, overpass (OP), SH, stadium (STA), ST, TC, train station (TS), vehicle (VE) and windmill (WM). In our experiments, we keep the image size at the original size of $800\times800$ pixels during both training and inference.

HRSC2016\cite{liu2017high} is a challenging high resolution ship detection dataset, which contains 1,061 aerial images from two scenarios including inshore and offshore ships. The training, validation and test sets include 436 images, 181 images and 444 images, respectively. The size of images ranges from $300\times300$ to $1500\times900$ pixels. In our experiments, the images of HRSC2016 are resized to $800\times800$ pixels during both training and inference.

\subsubsection{Implementation Details}
Unless specified, we adopt ResNet-50\cite{he2016deep} pretrained on ImageNet\cite{deng2009imagenet} as the backbone network, and FPN\cite{lin2017feature} as the neck network in our experiments. Then, the convolution weight that composes the prediction head is initialized with normal distribution, where the standard deviation is set to be 0.01. Also, the convolution bias is enabled and initialized with 0. We use the group normalization\cite{wu2018group} in the network, which performs better than the batch normalization\cite{ioffe2015batch}. $\delta$ and $\gamma$ are set to be 0.25 and 2.0 respectively in focal loss\cite{retinanet}, and $\beta$ in the smooth L1 loss is set as 0.01.

During training, we use stochastic gradient descent (SGD)\cite{bottou2010large} as the optimization approach unless specified, and the weight decay, momentum and gamma of SGD are set to be 0.0001, 0.9 and 0.1, respectively. In addition, to stabilize the training process and yield better performance, we adopt a learning rate scheduler with the combination of warm-up training and multi-step learning rate. Here, a linear warm-up method is applied, where the warm-up factor is set as 0.001 with 1000 iterations at the beginning of training, and random flipping is employed for data augmentation. All the experiments were implemented on a computer with four NVIDIA A100 GPUs, while the inference speed is measured on a single GeForce RTX 3090 GPU with 24GB memory. 

For experiments on DOTA dataset, our detectors are trained for 40k iterations with 16 images per device, and the learning rate is initialized with 0.045 but decayed to 1/10 at the iterations of 26k and 35k. As for DIOR-R dataset, models are trained for 20k iterations, and the learning rate is also initialized as 0.045 and decayed to 1/10  at 13k and 17.5k iterations. For experiments on HRSC2016 dataset, we train the models for 90k iterations, and the learning rate is initialized with 0.02 and decays at 60k and 80k iterations with a total of 16 images per batch.

\subsubsection{Evaluation Metrics}
For evaluating the accuracy of detectors, the mean Average Precision (mAP) with IoU threshold 0.5 is adopted, which is the widely used metric in object detection tasks. mAP$_{07}$ and mAP$_{12}$ indicate the mAP under PASCAL VOC2007 and VOC2012 metrics\cite{everingham2010pascal}. The Average Precision (AP) of each category is employed to validate the detection accuracy of different categories. The frames per second (fps) is used to evaluate the inference speed.

\subsection{Parameter Selection}

\begin{figure}[t]
    \centering
    \includegraphics[width=1.0\linewidth]{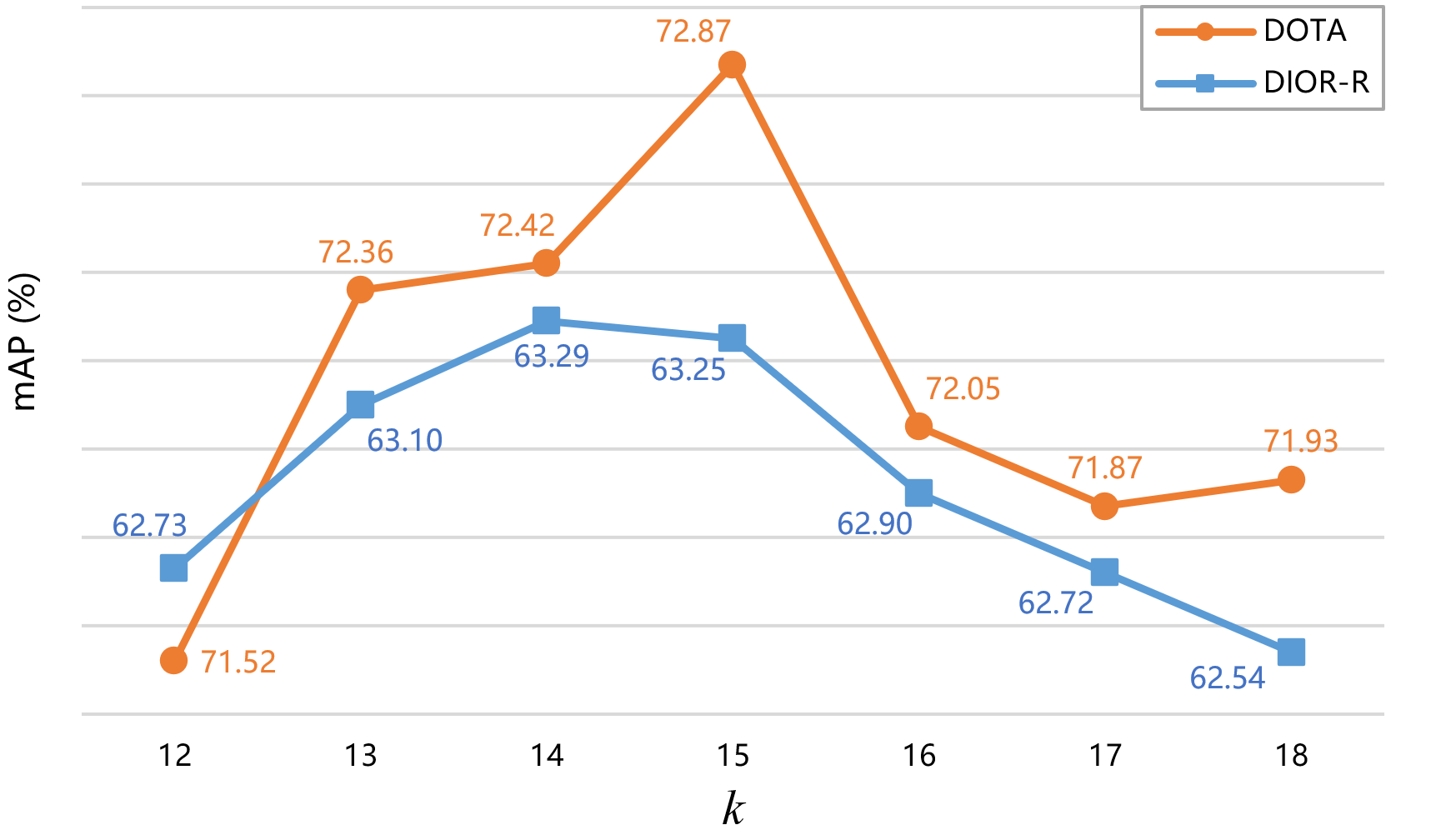}
    \caption{Performance evaluation in terms of mAP with different sample number $k$ on two datasets. The orange curve represents mAP on DOTA dataset, and the blue curve represents mAP on DIOR-R dataset.}
    \label{fig:parameter-k}
\end{figure}

The sample number $k$ in ADS strategy is empirically selected in our study. Due to its importance, we conducted experiments on two complex datasets, i.e., DOTA and DIOR-R, to quantitatively analyze the impact of the choice of $k$ on detection performance. In addition, we found that the number of samples affects the training duration to some extent, so for the sake of efficiency, we test $k$ from 12 to 18, and the results are given in Fig. \ref{fig:parameter-k}.
\begin{table*}[hbp]
    \centering
    \caption{Results of ablation study on  DOTA dataset. The best result is highlighted in bold.}
    \resizebox{\linewidth}{!}{
        \begin{tabular}{c|ccccccccccccccc|c}
        \Xhline{1pt}
        Methods & PL    & BD    & BR    & GTF   & SV    & LV    & SH    & TC    & BC    & ST    & SBF   & RA    & HA    & SP    & HC    & mAP (\%) \bigstrut\\
        \hline
        R-FCOS (Baseline) & 88.60  & 81.08 & 45.74 & 64.58 & 79.83 & 77.89 & 86.95 & 90.52 & 80.24 & 84.09 & 40.55 & 58.31 & 66.24 & 71.78 & 49.70  & 71.07 \bigstrut[t]\\
        +EARL (w/o -E, -W) & 89.87 & 80.57 & 46.22 & \textbf{65.61} & 80.51 & 78.35 & 87.39 & \textbf{90.76} & 78.41 & \textbf{86.28} & 45.10  & 61.42 & 65.08 & 69.56 & 51.70  & 71.79 (+0.72) \\
        +EARL (w/o -A, -W) & 89.12 & 77.67 & 46.98 & 65.06 & 80.27 & 78.03 & \textbf{87.68} & 90.54 & 82.65 & 85.23 & 47.67 & 60.58 & \textbf{67.08} & 70.98 & 44.64 & 71.61 (+0.54) \\
        +EARL (w/o -W) & \textbf{90.05} & \textbf{81.16} & 47.01 & 65.17 & 80.39 & 79.85 & 87.10  & 90.37 & \textbf{83.11} & 86.17 & 44.32 & 65.41 & 66.82 & 71.10  & 50.56 & 72.57 (+1.50) \\
        +EARL  & 89.76 & 78.79 & \textbf{47.01} & 65.20  & \textbf{80.98} & \textbf{79.99} & 87.33 & 90.74 & 79.17 & 86.23 & \textbf{49.09} & \textbf{65.87} & 65.75 & \textbf{71.86} & \textbf{55.21} & \textbf{72.87 (+1.80)} \bigstrut[b]\\
        \Xhline{1pt}
    \end{tabular}}
    \label{tab:1.1}
\end{table*}
\begin{table*}[hbp]
    \centering
    \caption{Results of ablation study on  DIOR-R dataset. The best result is highlighted in bold.}
    \resizebox{\linewidth}{!}{
        \begin{tabular}{c|cccccccccccccccccccc|c}
        \Xhline{1pt}
        Methods & APL & APO & BF & BC & BR & CH & ESA & ETS & DAM & GF & GTF & HA & OP & SH & STA & ST & TC & TS & VH & WM & mAP (\%) \bigstrut\\
        \hline
        R-FCOS (Baseline) & 56.19 & 37.61 & 76.82 & 85.71 & 26.82 & 74.57 & 74.45 & 59.26 & 20.24 & 77.66 & 77.74 & 39.60 & 46.08 & 81.08 & 69.56 & 63.34 & 85.53 & 53.91 & 39.81 & 65.71 & 60.58 \bigstrut[t]\\
        +EARL (w/o -E, -W) & 55.76 & 39.89 & 74.50 & 83.94 & \textbf{30.48} & 76.29 & 74.22 & \textbf{63.76} & 24.05 & 78.56 & 76.90 & 35.44 & \textbf{49.20} & 82.00 & 70.26 & 61.98 & 84.08 & 57.32 & 43.88 & \textbf{67.17} & 61.48 (+0.90) \\
        +EARL (w/o -A, -W) & \textbf{61.43} & 41.14 & 75.76 & \textbf{86.24} & 26.85 & 76.13 & 74.69 & 59.51 & 22.41 & 79.28 & 77.95 & 39.64 & 46.20 & 81.21 & 69.76 & \textbf{67.24} & 85.27 & 55.10 & 39.29 & 64.02 & 61.46 (+0.88) \\
        +EARL (w/o -W) & 58.38 & 43.95 & 75.59 & 83.95 & 29.99 & 76.91 & \textbf{76.89} & 61.08 & 22.39 & 79.31 & \textbf{79.32} & 38.75 & 48.26 & 82.37 & 70.54 & 66.23 & 86.34 & 54.51 & \textbf{44.00} & 64.75 & 62.18 (+1.60) \\
        +EARL  & 61.13 & \textbf{44.83 }& \textbf{77.74} & 84.66 & 30.42 & \textbf{78.50} & 76.73 & 62.79 & \textbf{24.67} & \textbf{79.58} & 79.19 & \textbf{39.64} & 48.18 & \textbf{82.64} & \textbf{76.59} & 64.90 & \textbf{86.87} & \textbf{57.98} & 43.82 & 64.90 & \textbf{63.29 (+2.71)} \bigstrut[b]\\
        \Xhline{1pt}
    \end{tabular}}
    \label{tab:1.2}
\end{table*}

For the performance on DOTA observed from Fig. \ref{fig:parameter-k},  when $k$=12, although the quality of each selected sample is guaranteed, the number of positive samples is inadequate for the model to learn feature distribution, so the mAP has only reached 71.52\%. The performance of the model gradually improves as $k$ increases, and the detector achieves the best mAP of 72.87\% when $k$=15. However, as $k$ continues to increase, more background noise are included since samples located near the boundaries of targets are selected, so the mAP score drops from 72.87\% to 71.93\% when $k$=18. 
Whereas for the selection of $k$ on DIOR-R, it shows the similar performance trend as that on DOTA, where the best performance is obtained when $k$ is set as 14. It can be seen that various $k$ selections yield relatively stable detection performance and can be simply changed the value of $k$ to suit different datasets.

\subsection{Ablation Study}
In this section, a series of ablation experiments were conducted on three popular datasets, i.e., DOTA, DIOR-R and HRSC2016, and the detailed results are listed in Tables \ref{tab:1.1}, \ref{tab:1.2} and \ref{tab:1.3}, respectively. In addition, we also visualized the gradient-based feature heatmaps\cite{chattopadhay2018grad} to further understand the effectiveness of each component of EARL. The visualization results are illustrated in Fig. \ref{fig:feature}. Here, the components of proposed EARL are indicated in abbreviated form, i.e., `-A' indicates ADS strategy, `-E' refers to DED strategy, and `-W' means SDW module. In addition, `w/o' is applied to indicate without specified modules.

\subsubsection{Effectiveness of ADS}

The proposed ADS strategy resolves the insufficient and imbalanced sampling problem for targets with extreme scales and aspect ratios. As shown in rows 1 and 2 of Tables \ref{tab:1.1}, \ref{tab:1.2} and \ref{tab:1.3}, the detectors trained with our ADS achieve the improvement of 0.72\% and 0.90\% in terms of mAP on DOTA and DIOR-R, and give a significant improvement of 2.28\% in terms of mAP$_{12}$ on HRSC2016, respectively, which demonstrates the effectiveness and robustness of ADS. In addition, the proposed ADS strategy boosts the AP performance of BR, SV, GTF and SBF on DOTA dataset by 0.48\%, 0.68\%, 1.03\% and 4.55\% compared with the baseline method, respectively, which indicates that our ADS is beneficial in assigning samples to targets with extreme scales and aspect ratios.

\begin{figure}[t]
    \centering
    \includegraphics[width=1\linewidth]{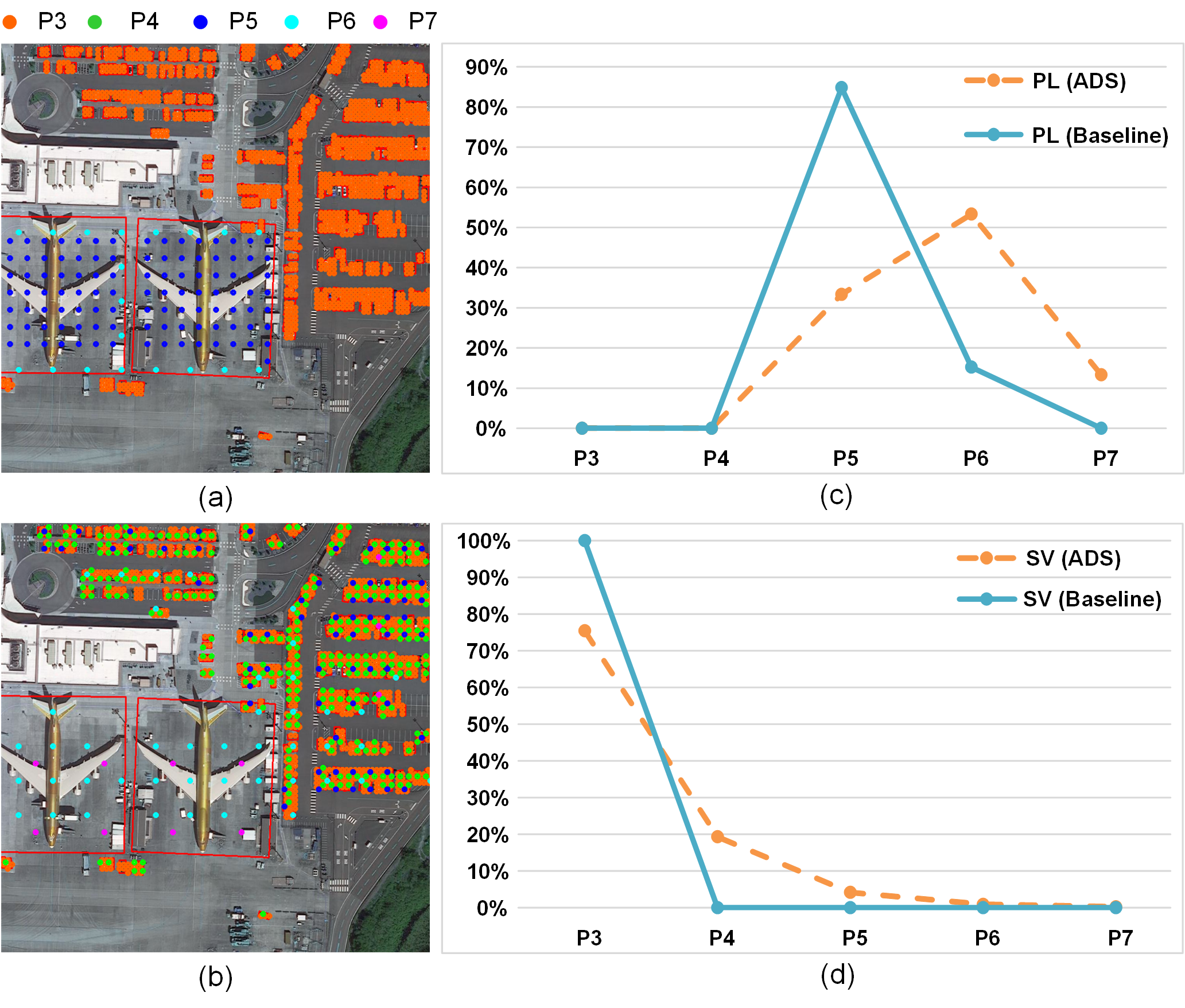}
    \caption{Comparison of sample numbers of PL and SV on each level of feature maps, where (a) and (b) are the visualization of sampling results, (c) and (d) are the statistics of the percent of samples on each level for PL and SV, respectively. Best viewed in colour.}
    \label{fig:effectADS}
\end{figure}

Taking PL and SV for instance, when using the baseline method, samples are restricted to be assigned on specific level of feature maps, as illustrated in Fig.~\ref{fig:effectADS}, and yields feature insufficient for various scales of targets. By employing ADS strategy, samples are selected more balanced for both PL and SV on each feature map, hence features with more various scales can be learned. However, although ADS strategy is able to obtain a better scale-level distribution, it introduces background noise in higher-level feature maps for specific categories, e.g., BD and BC, and brings down the accuracy. In order to mitigate the noise, DED with SDW are applied and discussed in the following subsections.

\begin{table}[htbp]
    \centering
    \caption{Results of ablation study on HRSC2016 dataset. The best result is highlighted in bold.}
    \begin{tabular}{c|cccccccccccccccccccc|c}
        \Xhline{1pt}
        Methods & mAP$_{07}$ (\%) & mAP$_{12}$ (\%) \bigstrut\\
        \hline
        R-FCOS (Baseline)  & 87.16         & 89.79 \bigstrut[t]\\
        +EARL (w/o -E, -W) & 87.97 (+0.81) & 92.07 (+2.28)\\
        +EARL (w/o -A, -W) & 87.86 (+0.70) & 90.72 (+0.93) \\
        +EARL (w/o -W)     & 88.75 (+1.59) & 92.78 (+2.99) \\
        +EARL              & \textbf{88.98 (+1.82)} & \textbf{93.04 (+3.25)} \bigstrut[b]\\
        \Xhline{1pt}
    \end{tabular}
    \label{tab:1.3}
\end{table}

\subsubsection{Effectiveness of DED}
The DED strategy is proposed to make the sample distribution more flexible to fit the shapes and orientations of targets. As shown in rows 1 and 3 of Tables \ref{tab:1.1}, \ref{tab:1.2} and \ref{tab:1.3}, our DED achieves 0.54\% and 0.88\% mAP improvement by replacing the original spatial assignment in the baseline method, i.e., R-FCOS, on DOTA and DIOR-R, respectively. Besides, it also provides 0.70\% mAP$_{07}$ and 0.93\% mAP$_{12}$ improvement on HRSC2016. The results demonstrate the effectiveness of our DED strategy. In addition, from rows 1 and 3 of Table~\ref{tab:1.1}, the AP of the elongated targets, such as BR and HA, are improved by 1.24\% and 0.84\%, respectively, which further proves that our DED is more suitable for detecting targets with large aspect ratios. 

Moreover, when combining DED to the proposed ADS, the mAP is improved by 1.50\% and 1.60\% on DOTA and DIOR-R, and the mAP$_{07}$ and mAP$_{12}$ are improved by 1.59\% and 2.99\% on HRSC2016, respectively, compared with the baseline method, as shown in rows 1 and 4 of Tables~\ref{tab:1.1}, \ref{tab:1.2} and \ref{tab:1.3}. The results further demonstrate that our DED strategy can further unlock the potential of ADS strategy, due to its ability to filter out low-quality samples.

\subsubsection{Effectiveness of SDW}
\begin{figure}[t]
    \centering
    \includegraphics[width=0.98\linewidth]{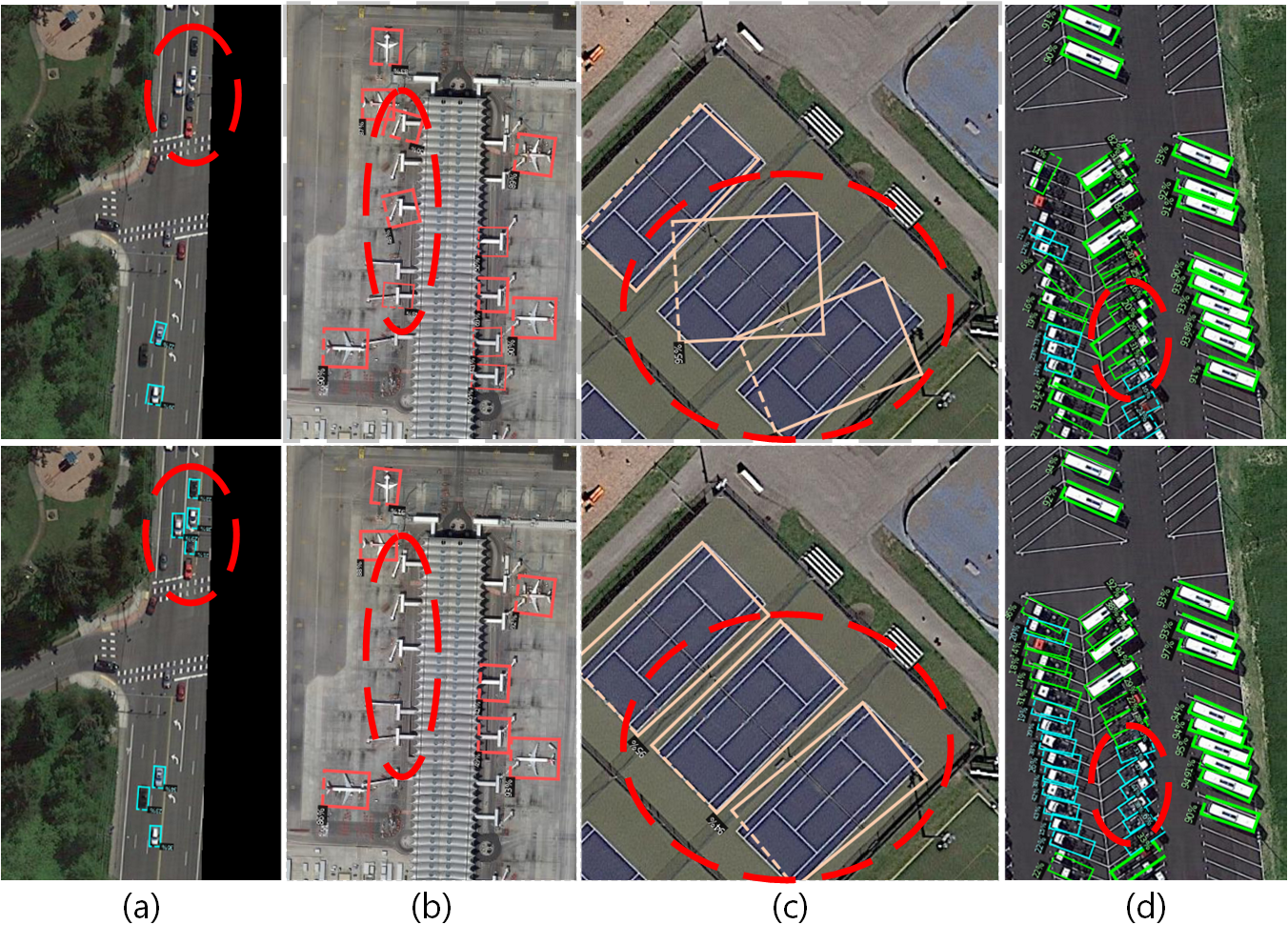}
    \caption{Visualization of the detection results with and without the SDW module on DOTA dataset. The first and second rows represent the results of without and with SDW module, respectively. Our method with SDW shows the improved  overall detection performance, i.e., (a) Reducing the missed targets; (b) Reducing false alarms; (c) Better regression;  (d) Better classification.}
    \label{fig:effectSDW}
\end{figure}

SDW module is proposed to further mitigate the influence from low-quality samples by weighting the samples adaptively, and achieve high-quality sampling, which yields the mAP improvement of 0.30\% and 1.11\% on DOTA and DIOR-R, and gives the mAP$_{12}$ improvement of 0.26\% on HRSC2016, respectively, as compared rows 4 and 5 of Tables \ref{tab:1.1}, \ref{tab:1.2} and \ref{tab:1.3}. In addition, SDW module significantly improves AP for difficult categories, e.g., SBF, HC and DAM with 4.77\%, 4.65\% and 2.28\% improvement respectively, by learning feature information from higher-quality samples. Moreover, to further demonstrate the effect of SDW module, the visualization comparison is provided to show the detection performance of our method with and without SDW module, as illustrated in Fig.~\ref{fig:effectSDW}, where we can see that our SDW module can further enhance the overall detection performance.

Overall, by employing ADS, DED and SDW, the proposed EARL strategy significantly enhances the detection performance in RSIs with 1.80\% and 2.71\% mAP improvement on DOTA and DIOR-R, and 1.82\% mAP$_{07}$ and 3.25\% mAP$_{12}$ improvement on HRSC2016, respectively, as shown in rows 1 and 5 of Tables \ref{tab:1.1}, \ref{tab:1.2} and \ref{tab:1.3}. The results demonstrate the effectiveness and robustness of the proposed EARL strategy by taking into account of the characteristics of targets.

\subsubsection{Visualization Explanation}
\begin{figure}[tbp]
    \centering
    \includegraphics[width=1\linewidth]{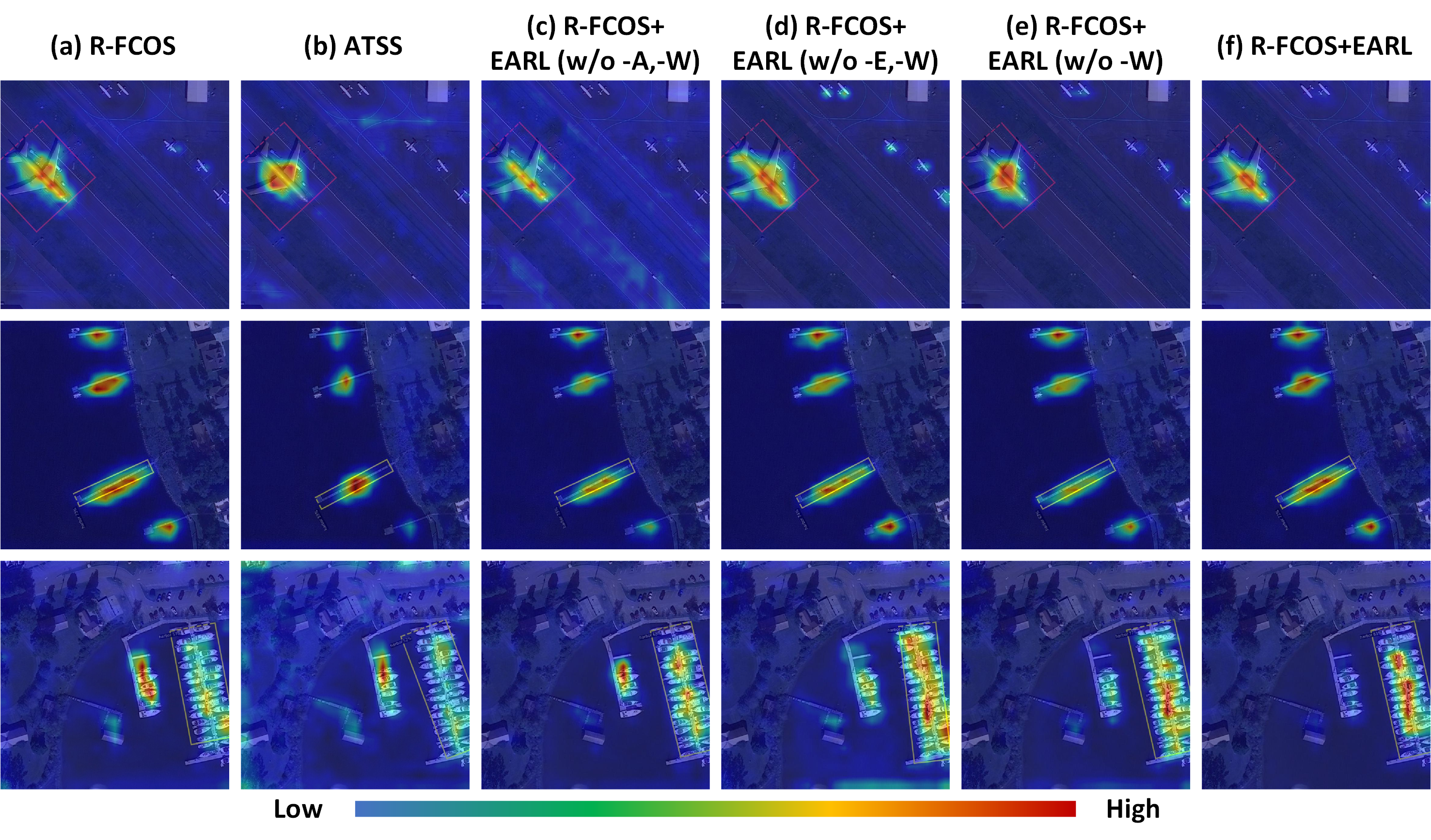}
    \caption{Gradient-based class-discriminative saliency maps visualization comparison of the detectors trained with different label assignment strategies on DOTA dataset. Best viewed in colour.}
    \label{fig:feature}
\end{figure}
To further understand the effectiveness of each proposed component of EARL, the gradient-based class-discriminative saliency maps\cite{chattopadhay2018grad} are displayed to visualize the feature heatmaps of detectors with different label assignment strategies, as illustrated in Fig.~\ref{fig:feature}.
As illustrated in row 1 of Fig.~\ref{fig:feature} (a) and (c), EARL (w/o -A, -W) constricts the sampling range based on R-FCOS, so that less attention is paid to the background. From rows 1 and 2 of Fig.~\ref{fig:feature} (a) and (b), ATSS focuses too much on the central region resulting in its failure to learn sufficient features and can hardly reflect the orientation and shape characteristics. Row 3 of Fig.~\ref{fig:feature} (a) and (d) show that EARL (w/o -E, -W) can extract richer information of targets with extreme scales and aspect ratios. However, due to its ability to obtain training samples from multi-level feature maps, it may introduce background noise. As shown in Fig.~\ref{fig:feature} (d) and (e), using DED strategy can mitigate the impact of noise and release the potential of ADS strategy. In addition, as illustrated in Fig.~\ref{fig:feature} (e) and (f), SDW module makes the detector more focused on the high-quality samples to learn better feature representation of the targets, thereby it can accurately focus on the target regions in the images.

\subsection{Comparison with Existing Label Assignment Strategies}
\begin{table}[t]
  \centering
  \caption{Comparison with different scale assignment strategies applied to R-FCOS. The best result is highlighted in bold.}
  \resizebox{\linewidth}{!}{
    \begin{tabular}{c|c|ccccc}
    \Xhline{1pt}
    \multicolumn{2}{c|}{Scale Assignment} & FSA\cite{huang2022general} & ALS\cite{yu2022object} & TKL\cite{zhang2020bridging} & TKT\cite{li2022oriented} & ADS \bigstrut\\
    \hline
    \multirow{2}[2]{*}{mAP (\%)} & DOTA & 71.07 & 56.12 & 70.14 & 68.67 & \textbf{71.79}\bigstrut[t]\\
      & DIOR-R & 60.58 & 41.91 & 57.63 & 52.52 & \textbf{61.48}\bigstrut[b]\\
    \Xhline{1pt}
    \end{tabular}}
  \label{tab:effectADS}
\end{table}
To further justify the effectiveness and superiority of the proposed scale assignment strategy, i.e., ADS, and the spatial assignment strategy, i.e., DED, a series of experiments were conducted on DOTA and DIOR-R datasets. For fair comparison, all compared label assignment strategies are evaluated on the same baseline detector (i.e., R-FCOS), and the results are listed in Tables \ref{tab:effectADS}, \ref{tab:effectDED} and \ref{tab:parameter-xi}.

\subsubsection{Comparison with Existing Scale Assignments}

The proposed ADS strategy is compared with four different scale assignment strategies, including the fixed scale assignment (FSA) strategy adopted in \cite{huang2022general,tian2019fcos,cheng2022anchor}, selecting all samples among all levels of feature maps (ALS) used in \cite{yu2022object}, selecting top-$k$ samples per feature level (TKL) utilized in \cite{zhang2020bridging}, and selecting top-$k$ samples for each target without sequential selection (TKT) used in \cite{li2022oriented}. The results are given in Table~\ref{tab:effectADS}.

It can be observed that using the ALS strategy leads to performance degradation due to the introduction of large amounts of low-quality samples and ambiguous samples. Regarding the TKL and TKT strategies, they cannot reasonably exploit the large scale variations of targets, leading to the inferior performance than our proposed ADS. Where our proposed ADS achieves the mAP of 71.79\% and 61.48\% on DOTA and DIOR-R datasets, respectively, which outperform other scale assignment strategies. 
The results demonstrate the effectiveness of our ADS strategy by considering the large variations in scales and aspect ratios of targets in RSIs, which achieves more flexible and adaptive scale assignment, and is more suitable for object detection in RSIs.

\begin{table}[t]
  \centering
  \caption{Comparison with different sampling range applied to R-FCOS. The best result is highlighted in bold.}
    \begin{tabular}{c|c|ccc}
    \Xhline{1pt}
    \multicolumn{2}{c|}{Sampling Range} & RBB & CA & DED \bigstrut\\
    \hline
    \multirow{2}[2]{*}{mAP (\%)} & DOTA & 71.07 & 70.61 & \textbf{71.61}\bigstrut[t]\\
      & DIOR-R & 60.58 & 60.61 & \textbf{61.46}\bigstrut[b]\\
    \Xhline{1pt}
    \end{tabular}
  \label{tab:effectDED}
\end{table}
\begin{figure}[t]
    \centering
    \includegraphics[width=1.0\linewidth]{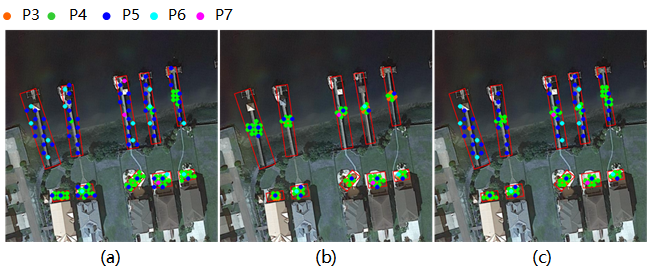}
    \caption{Illustration of the positive samples selected by obeying different sampling range, i.e., (a) RBB, (b) CA, and (c) DED. Best viewed in colour.}
    \label{fig:exp_ed}
\end{figure}

\subsubsection{Comparison with Existing Spatial Assignments}
The proposed DED strategy is compared with other sampling range for spatial assignment based on the R-FCOS, i.e., the rectangle bounding box (RBB) and the central area (CA). The results are provided in Table~\ref{tab:effectDED}.

From Table~\ref{tab:effectDED},  we can see that our DED strategy improves mAP by 0.54\% and 0.88\% when compared with RBB and CA, respectively. 
The reason is that low-quality samples are often involved in RBB, as illustrated in Fig.~\ref{fig:exp_ed} (a), and using CA often directly discards foreground samples at both ends of bounding box, resulting in insufficient sampling, as shown in Fig.~\ref{fig:exp_ed} (b). 
When employing the proposed DED method, the sample distribution is more flexible to fit the shapes and orientations of targets, hence is more suitable for oriented object detection in RSIs, and achieves the mAP of 71.61\% and 61.46\% on DOTA and DIOR-R datasets, respectively. 

In addition, the dynamic elliptical distribution in our proposed DED is compared with those with fixed ratio factor $\xi$, which is introduced to control the size of sampling range.
The results are given in Table \ref{tab:parameter-xi}, where the fixed elliptical distribution achieves their peak performance of 72.49\% and 62.97\% mAP when $\xi$=0.6 on DOTA and DIOR-R datasets, respectively. Whereas our dynamic elliptical distribution achieves better performance of 72.87\% and 63.29\% on DOTA and DIOR-R datasets by dynamically adjusting sampling range according to the aspect ratios of targets, and avoids parameter selection.
\begin{table}[htbp]
  \centering
  \caption{Comparison with different elliptical distribution with various ratio factor $\xi$ applied to R-FCOS. The best result is highlighted in bold.}
    \begin{tabular}{c|c|ccccc}
    \Xhline{1pt}
    \multicolumn{2}{c|}{$\xi$} & 0.4 & 0.6 & 0.8 & 1.0 & Ours \bigstrut\\
    \hline
    \multirow{2}[2]{*}{mAP (\%)} & DOTA & 71.99 & 72.49 & 72.16 & 71.76 & \textbf{72.87}\bigstrut[t]\\
      & DIOR-R & 62.38 & 62.97 & 62.68 & 62.36 & \textbf{63.29}\bigstrut[b]\\
    \Xhline{1pt}
    \end{tabular}
  \label{tab:parameter-xi}
\end{table}

\begin{table*}[!t]
    \centering
    \caption{Performance comparison with state-of-the-art methods on DOTA dataset. $\ddagger$ indicates multi-scale training and testing. $\dagger$ means our re-implementation with the official code. The best result is highlighted in bold.}
    \resizebox{\linewidth}{!}{
    \begin{tabular}{c|c|ccccccccccccccc|cc}
    \Xhline{1pt}
    Methods & Backbone & PL & BD & BR & GTF & SV & LV & SH & TC & BC & ST & SBF & RA & HA & SP & HC & mAP (\%) & fps\\
    \hline
    \textit{\textbf{Anchor-based}}\\
    \hline
    CenterMap-Net\cite{wang2020learning} & Res50 & 89.02 & 80.56 & 49.41 & 61.98 & 77.99 & 74.19 & 83.74 & 89.44 & 78.01 & 83.52 & 47.64 & 65.93 & 63.68 & 67.07 & 61.59 & 71.59 & - \\
    R$^3$Det\cite{yang2019r3det} & Res50 & 89.02 & 75.47 & 43.86 & 65.84 & 75.83 & 73.44 & 86.03 & 90.57 & 81.11 & 82.84 & 55.57 & 59.10 & 56.57 & 70.31 & 50.45 & 70.40 & - \\
    S$^2$A-Net\cite{han2021align}  & Res50 & 89.11 & 82.84 & 48.37 & 71.11 & 78.11 & 78.39 & 87.25 & 90.83 & 84.90 & 85.64 & 60.36 & 62.60 & 65.26 & 69.13 & 57.94 & 74.12 & 17.6 \\
    S$^2$A-Net\cite{han2021align}$^{\ddagger}$  & Res50 & 88.89 & 83.60 & \textbf{57.74} & \textbf{81.95} & 79.94 & 83.19 & \textbf{89.11} & 90.78 & 84.87 & 87.81 & \textbf{70.30} & 68.25 & \textbf{78.30} & 77.01 & 69.58 & 79.42 & 17.6 \\
    RoI-Trans.\cite{ding2019learning}$^{\ddagger}$ & Res101 & 88.53 & 77.91 & 37.63 & 74.08 & 66.53 & 62.97 & 66.57 & 90.50 & 79.46 & 76.75 & 59.04 & 56.73 & 62.54 & 61.29 & 55.56 & 67.74 & 7.8 \\
    SCRDet\cite{yang2019scrdet}$^{\ddagger}$ & Res101 & 89.98 & 80.65 & 52.09 & 68.36 & 68.36 & 60.32 & 72.41 & 90.85 & 87.94 & 86.86 & 65.02 & 66.68 & 66.25 & 68.24 & 65.21 & 72.61 & 9.5 \\
    Gliding-Ver.\cite{xu2020gliding}$^{\ddagger}$ & Res101 & 89.64 & 85.00 & 52.26 & 77.34 & 73.01 & 73.14 & 86.82 & 90.74 & 79.02 & 86.81 & 59.55 & 70.91 & 72.94 & 70.86 & 57.32 & 75.02 & 13.1 \\
    Hou \cite{hou2022refined} & Res101 & 89.32 & 76.05 & 50.33 & 70.25 & 76.44 & 79.45 & 86.02 & 90.84 & 82.80 & 82.50 & 58.17 & 62.46 & 67.38 & 71.93 & 45.52 & 72.63 & - \\
    Hou \cite{hou2022refined}$^{\ddagger}$ & Res101 & 88.69 & 79.41 & 52.26 & 65.51 & 74.72 & 80.83 & 87.42 & 90.77 & 84.31 & 83.36 & 62.64 & 58.14 & 66.95 & 72.32 & 69.34 & 74.44 & - \\
    \hline 
    \textit{\textbf{Anchor-free}}\\
    \hline 
    ATSS\cite{zhang2020bridging}$^{\dagger}$ & Res50 & 88.47 & 80.05 & 47.27 & 60.65 & 79.85 & 78.80 & 87.41 & 90.75 & 77.37 & 85.24 & 43.22 & 60.80 & 66.52 & 70.95 & 41.82 & 70.61 & - \\
    SASM\cite{hou2022shape} & Res50 & 86.42 & 78.97 & 52.47 & 69.84 & 77.30 & 75.99 & 86.72 & 90.89 & 82.63 & 85.66 & 60.13 & 68.25 & 73.98 & 72.22 & 62.37 & 74.92 & - \\
    AOPG\cite{cheng2022anchor} & Res50 & 89.27 & 83.49 & 52.50 & 69.97 & 73.51 & 82.31 & 87.95 & 90.89 & 87.64 & 84.71 & 60.01 & 66.12 & 74.19 & 68.30 & 57.80 & 75.24 & 12.4 \\
    Oriented RepPoints\cite{li2022oriented} & Res50 & 87.02 & 83.17 & 54.13 & 71.16 & 80.18 & 78.40 & 87.28 & \textbf{90.90} & 85.97 & 86.25 & 59.90 & 70.49 & 73.53 & 72.27 & 58.97 & 75.97 & - \\
    FSDet\cite{yu2022object} & Res50 & 89.20 & 84.33 & 55.38 & 74.81 & 80.85 & 78.79 & 88.24 & 90.85 & 86.85 & 85.85 & 67.14 & 64.79 & 76.49 & 75.11 & 69.02 & 77.85 & - \\
    GGHL\cite{huang2022general} & DarkNet53 & 89.74 & \textbf{85.63} & 44.50 & 77.48 & 76.72 & 80.45 & 86.16 & 90.83 & \textbf{88.18} & 86.25 & 67.07 & 69.40 & 73.38 & 68.45 & 70.14 & 76.95 & \textbf{42.3} \\
    IENet\cite{lin2019ienet} & Res101 & 88.15    & 71.38    & 34.26    & 51.78    & 63.78    & 65.63    & 71.61    & 90.11    & 71.07    & 73.63    & 37.62    & 41.52    & 48.07    & 60.53    & 49.53    & 61.24  & 16.9 \\
    Axis-Learning\cite{xiao2020axis} & Res101 & 79.53 & 77.15 & 38.59 & 61.15 & 67.53 & 70.49 & 76.30 & 89.66 & 79.07 & 83.53 & 47.27 & 61.01 & 56.28 & 66.06 & 36.05 & 65.98 & 14.1 \\
    BBAVectors\cite{yi2021oriented} & Res101 & 88.35 & 79.96 & 50.69 & 62.18 & 78.43 & 78.98 & 87.94 & 90.85 & 83.58 & 84.35 & 54.13 & 60.24 & 65.22 & 64.28 & 55.70 & 72.32 & 18.4 \\
    \hline
    R-FCOS & Res50 & 88.60  & 81.08 & 45.74 & 64.58 & 79.83 & 77.89 & 86.95 & 90.52 & 80.24 & 84.09 & 40.55 & 58.31 & 66.24 & 71.78 & 49.70  & 71.07 & 33.5 \\
    \textbf{R-FCOS-EARL} & Res50 & 89.76 & 78.79 & 47.01 & 65.20 & 80.98 & 79.99 & 87.33 & 90.74 & 79.17 & 86.23 & 49.09 & 65.87 & 65.75 & 71.86 & 55.21 & 72.87 & 33.5 \\
    \textbf{R-FCOS-EARL}$^{\ddagger}$ & Res50 & \textbf{90.13} & 83.90 & 47.19 & 72.17 & \textbf{81.54} & 84.26 & 88.24 & 90.69 & 79.10 & 86.71 & 60.47 & 72.21 & 71.26 & 73.81 & 58.80 & 76.03 & 33.5 \\
    \textbf{R-FCOS-EARL} & Res101 & 89.79 & 84.05 & 48.69 & 65.89 & 80.39 & 78.38 & 87.22 & 90.46 & 79.90 & 86.14 & 47.01 & 66.44 & 66.79 & 72.42 & 49.81 & 72.89 & 25.9 \\
    \textbf{R-FCOS-EARL}$^{\ddagger}$ & Res101 & 89.28 & 83.80 & 49.34 & 72.27 & 80.69 & 84.01 & 88.30 & 90.77 & 82.35 & 87.39 & 59.90 & \textbf{75.59} & 69.61 & 71.93 & 53.82 & 75.94 & 25.9 \\
    \hline
    RTMDet-R\cite{lyu2022rtmdet}$^{\dagger}$ & CSPNeXt-L & 89.44 & 77.91 & 54.97 & 74.54 & 80.05 & 83.15 & 89.06 & \textbf{90.90} & 83.75 & 86.32 & 60.06 & 68.53 & 78.03 & 79.74 & 64.78 & 77.41 & 29.3 \\
    \textbf{RTMDet-R-EARL} & CSPNeXt-L & 88.38 & 77.93 & 52.71 & 74.55 & 80.29 & 82.88 & 88.76 & 90.58 & 83.34 & 86.85 & 62.38 & 68.39 & 75.85 & 80.54 & 69.61 & 77.54 & 29.3 \\
    \textbf{RTMDet-R-EARL}$^{\ddagger}$ & CSPNeXt-L & 87.92 & 85.62 & 56.48 & 80.47 & 80.74 & \textbf{84.82} & 88.63 & 90.74 & 86.16 & \textbf{88.05} & 69.76 & 70.12 & 78.06 & \textbf{82.34} & \textbf{75.31} & \textbf{80.35} & 29.3 \\
    \Xhline{1pt}
    \end{tabular}}
    \label{tab:5}
\end{table*}
\begin{figure*}[!t]
    \centering
    \includegraphics[width=0.99\linewidth]{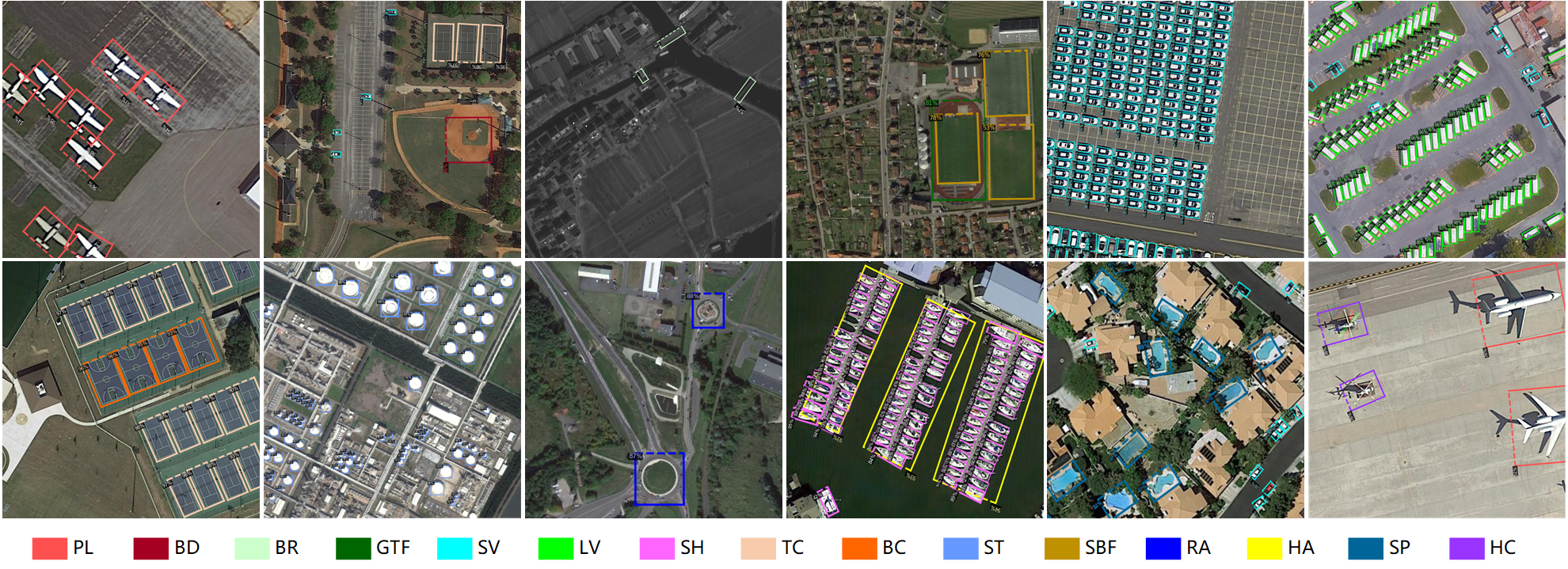}
    \caption{Visualization of detection results on  DOTA dataset with our method. Best viewed in colour.}
    \label{fig:visualDOTA}
\end{figure*}
\begin{figure*}[!t]
    \centering
     \includegraphics[width=1.0\linewidth]{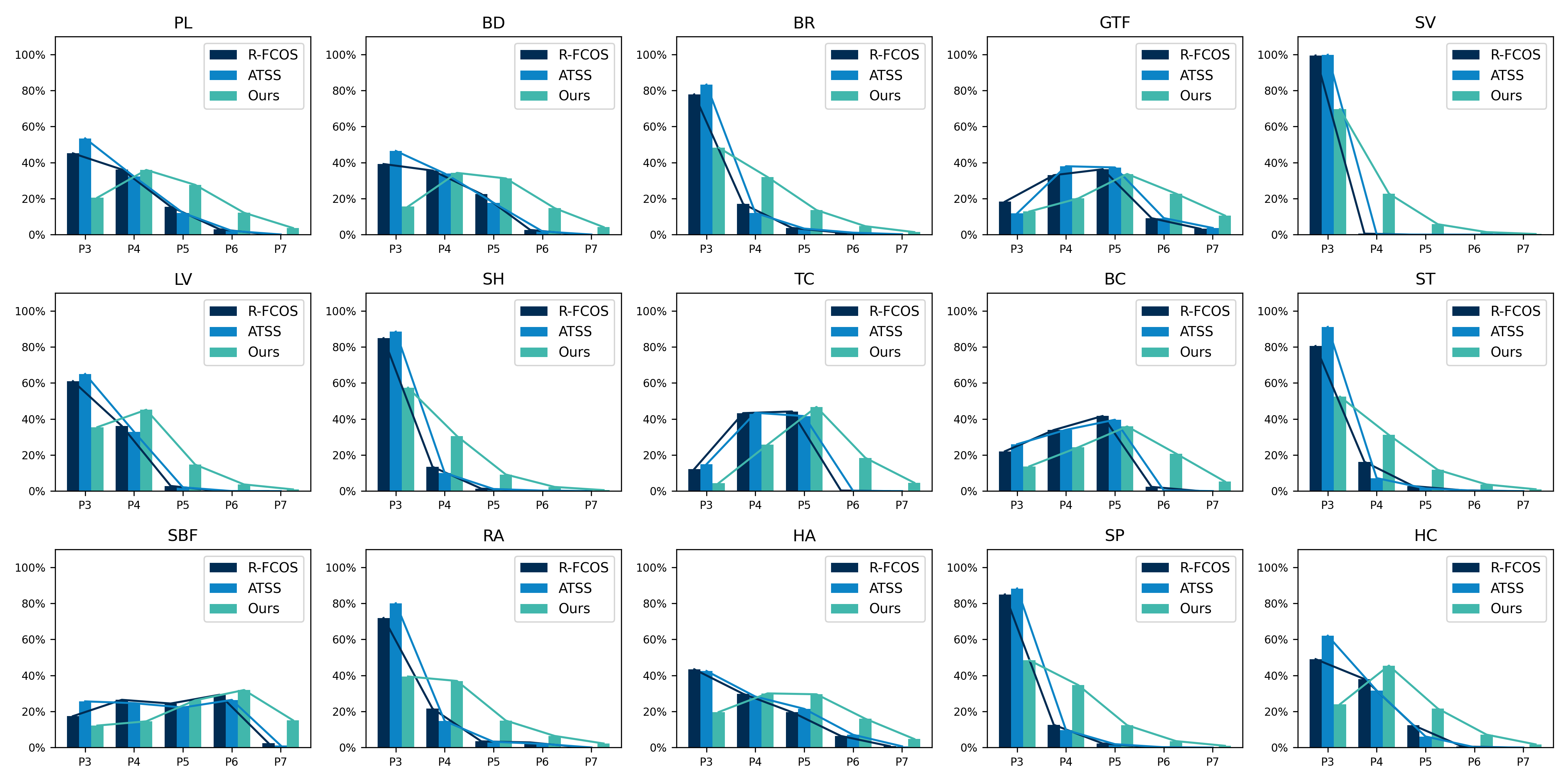}
     \caption{Histogram of the percentage of the number of positive samples selected at different feature levels using different label assignment strategies on  DOTA dataset. Best viewed in colour.}
     \label{fig:exp_ads}
\end{figure*}
\subsection{Comparison with State-of-the-Art}
To evaluate the performance of the proposed EARL strategy, we conducted experiments compared with the state-of-the-art methods on three public RSIs datasets, i.e., DOTA, DIOR-R and HRSC2016, and the results are listed in Tables \ref{tab:5}, \ref{tab:dior-r} and \ref{tab:hrsc2016}, respectively. Note that, for the purpose of fair comparison, these results are from the corresponding papers of the comparison methods and datasets, unless specified.
In addition, to fully validate the effectiveness and robustness of the proposed EARL strategy, a simple detector, i.e., R-FCOS, and an advanced detector, i.e., RTMDet-R, are selected as the baseline methods, where our EARL is deployed with them, denoted as R-FCOS-EARL and RTMDet-R-EARL, respectively.

\subsubsection{Performance Evaluation on DOTA Dataset}
Table \ref{tab:5} gives the comparison of the proposed method with both anchor-based and anchor-free state-of-the-art detectors on DOTA dataset. It can be observed that anchor-based detectors can usually achieve higher performance, e.g., S$^2$A-Net achieves 79.42\% mAP. However, they often have limitations in inference speed due to complex structures and IoU calculations, which results in lower fps, e.g., 17.6 fps by S$^2$A-Net. As a balance of trade-off of accuracy and efficiency, when using ResNet50 as the backbone network, R-FCOS-EARL achieves mAP of 72.87\% and inference speed of 33.5 fps, even without any tricks.

When compared with anchor-free detectors, our R-FCOS-EARL method provides the promising performance of 72.87\% mAP with ResNet50 and 72.89\% mAP with ResNet101. Specifically, when applying multi-scale augmentation in experiments, our method achieves the mAP of 76.03\% with ResNet50 and 75.94\% with ResNet101 and obtains the best AP results for small targets, such as SV of 81.54\%, which further demonstrates the superiority of our method for detecting targets with extreme scales. 
The visualization of detection results with our method is displayed in Fig.~\ref{fig:visualDOTA}. It can be observed that the detector with our EARL strategy can accurately detect densely arranged targets and extreme targets benefited from the sufficient and balanced sampling.
Furthermore, our EARL is simply a training strategy which can be easily deployed to other detectors, e.g., when deployed on RTMDet-R, it obtains the state-of-the-art performance, i.e., 80.35\% mAP.

In addition, to further illustrate that our method can obtain a better scale-level distribution, as shown in Fig.~\ref{fig:exp_ads}, the proposed R-FCOS-EARL is compared with the most relevant label assignment strategy, ATSS, together with
the baseline method R-FCOS on DOTA dataset, by comparing the distribution of the positive samples among multi-level feature maps statistically. It can be seen that R-FCOS and ATSS achieve scale assignment according to the predefined scale constraints, which causes imbalanced sample distribution, and leads to the scale-level bias, especially for targets with extreme scales and aspect ratios, such as GTF and SV. Whereas our method with ADS strategy ensures that targets with different scales and shapes can assign training samples from all feature maps adaptively, which alleviates the imbalanced sampling thus improves detection performance. As a result, R-FCOS-EARL achieves the mAP of 72.87\%, which significantly improves mAP by 1.80\% and 2.26\% over R-FCOS and ATSS, respectively, as shown in Table \ref{tab:5}. Furthermore, the AP performance in Table \ref{tab:5} also confirms the improvement of our method for both extremely large and small targets, such as GTF, SBF, SV and LV. The above analysis proves that our EARL strategy is advantageous for object detection in RSIs.
\begin{table*}[t]
  \scriptsize
  \centering
  \caption{Performance comparison with state-of-the-art methods on  DIOR-R dataset. $\dagger$ means our re-implementation with the official code. The best result is highlighted in bold.}
  \resizebox{\linewidth}{!}{
    \begin{tabular}{c|c|cccccccccccccccccccc|c}
    \Xhline{1pt}
    Methods & Backbone & APL & APO & BF & BC & BR & CH & ESA & ETS & DAM & GF & GTF & HA & OP & SH & STA & ST & TC & TS & VH & WM & mAP (\%) \bigstrut\\
    \hline
    \textit{\textbf{Anchor-based}}\\
    \hline
    Gliding-Ver.\cite{xu2020gliding}  & Res50 & 65.4  & 28.9  & 75.0  & 81.3  & 33.9  & 74.3  & 64.7  & 70.7  & 19.6  & 72.3  & 78.7  & 37.2  & 49.6  & 80.2  & 69.3  & 61.1  & 81.5  & 44.8  & 47.7  & 65.0  & 60.1 \bigstrut[t]\\
    R$^3$Det\cite{yang2019r3det}      & Res50 & 62.6  & 43.4  & 71.7  & 81.5  & 36.5  & 72.6  & 79.5  & 64.4  & 27.0  & 77.4  & 77.2  & 40.5  & 53.3  & 79.7  & 69.2  & 61.1  & 81.5  & 52.2  & 43.6  & 64.1  & 61.9 \\
    S$^2$A-Net\cite{han2021align}      & Res50 & 65.4  & 42.0  & 75.2  & 83.9  & 36.0  & 72.6  & 75.1  & 65.1  & 28.0  & 75.6  & 80.5  & 35.9  & 52.1  & 82.3  & 65.9  & 66.1  & 84.6  & 54.1  & 48.0  & 69.7  & 62.9 \\
    RoI-Trans.\cite{ding2019learning} & Res50 & 63.3  & 37.9  & 71.8  & 87.5  & 40.7  & 72.6  & 68.1  & 78.7  & 26.9  & 69.0  & \textbf{82.7}  & 47.7  & 55.6  & 81.2  & 78.2  & 70.3  & 81.6  & 54.9  & 43.3  & 65.5  & 63.9  \\
    DODet\cite{cheng2022dual}         & Res50 & 63.4  & 43.4  & 72.1  & 81.3  & 43.1  & 72.6  & 70.8  & \textbf{78.8}  & 33.3  & 74.2  & 75.5  & 48.0  & \textbf{59.3}  & 85.4  & 74.0  & 71.6  & 81.5  & 55.5  & 51.9  & 66.4  & 65.1 \bigstrut[b]\\
    \hline
    \textit{\textbf{Anchor-free}}\\
    \hline
    SASM\cite{hou2022shape}           & Res50 & 61.4  & 46.0  & 73.2  & 82.0  & 29.4  & 71.0  & 69.2  & 53.9  & 30.6  & 70.0  & 77.0  & 39.3  & 47.5  & 78.6  & 66.1  & 62.9  & 79.9  & 54.4  & 40.6  & 63.0  & 59.8  \bigstrut[t]\\
    AOPG\cite{cheng2022anchor}        & Res50 & 62.4 & 37.8 & 71.6 & \textbf{87.6} & 40.9 & 72.5 & 78.0 & 65.4 & 31.1 & 73.2 & 81.9 & 42.3 & 54.5 & 81.2 & 72.7 & 71.3 & 81.5 & \textbf{60.0} & 52.4 & 70.0 & 64.4 \bigstrut[b]\\
    \hline
    R-FCOS                            & Res50 & 56.2 & 37.6 & 76.8 & 85.7 & 26.8 & 74.6 & 74.5 & 59.3 & 20.2 & 77.7 & 77.7 & 39.6 & 46.1 & 81.1 & 69.6 & 63.3 & 85.5 & 53.9 & 39.8 & 65.7 & 60.6 \bigstrut[t]\\
    \textbf{R-FCOS-EARL}              & Res50 & 61.1 & 44.8 & \textbf{77.7} & 84.7 & 30.4 & 78.5 & 76.7 & 62.8 & 24.7 & \textbf{79.6} & 79.2 & 39.6 & 48.2 & 82.6 & \textbf{76.6} & 64.9 & \textbf{86.9} & 58.0 & 43.8 & 64.9 & 63.3 \bigstrut[b]\\
    \hline
    RTMDet-R\cite{lyu2022rtmdet}$^\dagger$ & CSPNeXt-L & \textbf{71.1} & \textbf{49.9} & 71.8 & 86.7 & 43.3 & \textbf{80.9} & 80.6 & 70.9 & \textbf{34.5} & 76.4 & 81.6 & \textbf{49.4} & 57.6 & 89.4 & 71.5 & 77.9 & 81.6 & 57.9 & 52.6 & 65.6 & 67.5 \bigstrut[t]\\
    \textbf{RTMDet-R-EARL}                 & CSPNeXt-L & 71.0  & 46.8  & 76.3  & 84.1  & \textbf{46.7}  & 80.5  & \textbf{84.4}  & 71.3  & 32.0  & 77.9  & 80.9  & 48.6  & 58.4  & \textbf{89.5}  & 72.4  & \textbf{78.9}  & 81.5  & 56.7  & \textbf{54.9}  & \textbf{71.5}  & \textbf{68.2}  \bigstrut[b]\\
    \Xhline{1pt}
    \end{tabular}}
  \label{tab:dior-r}
\end{table*}
\begin{table}[t]
  \scriptsize
  \centering
  \caption{Performance comparison with different state-of-the-art methods on  HRSC2016 dataset. HG104 means Hourglass 104 \cite{newell2016stacked}. $\dagger$ means our re-implementation with the official code. The best result is highlighted in bold.}
  \begin{tabular}{c|c|cc}
    \Xhline{1pt}
    Methods & Backbone & mAP$_{07}$ (\%) & mAP$_{12}$ (\%)\\
    \hline
    \textit{\textbf{Anchor-based}}\\
    \hline
    RoI-Trans.\cite{ding2019learning} & Res101 & 86.2 & - \\
    Gliding-Ver.\cite{xu2020gliding} & Res101 & 88.2 & - \\
    DAL\cite{ming2020dynamic} & Res50 & 88.6 & - \\
    CenterMap-Net\cite{wang2020learning} & Res50 & - & 92.8 \\
    \hline
    \textit{\textbf{Anchor-free}}\\
    \hline
    IENet\cite{lin2019ienet} & Res101 & 75.0 & - \\
    Axis-Learning\cite{xiao2020axis} & Res101 & 78.2 & - \\
    TOSO\cite{feng2020toso} & Res101 & 79.3 & - \\
    BBAVectors\cite{yi2021oriented} & Res101 & 88.6 & - \\
    DRN\cite{pan2020dynamic} & HG104 & - & 92.7 \\
    \hline
    R-FCOS & Res50 & 87.2 & 89.8 \\
    \textbf{R-FCOS-EARL} & Res50 & 89.0 & 93.0\\
    \textbf{R-FCOS-EARL} & Res101 & 88.8 & 92.1\\
    \hline
    RTMDet-R\cite{lyu2022rtmdet}$^\dagger$ & CSPNeXt-L & 90.1 & 97.3 \\
    \textbf{RTMDet-R-EARL} & CSPNeXt-L & \textbf{90.3} & \textbf{98.0} \\
    \Xhline{1pt}
    \end{tabular}
  \label{tab:hrsc2016}
\end{table}
\begin{figure}[t]
    \centering
    \includegraphics[width=1.0\linewidth]{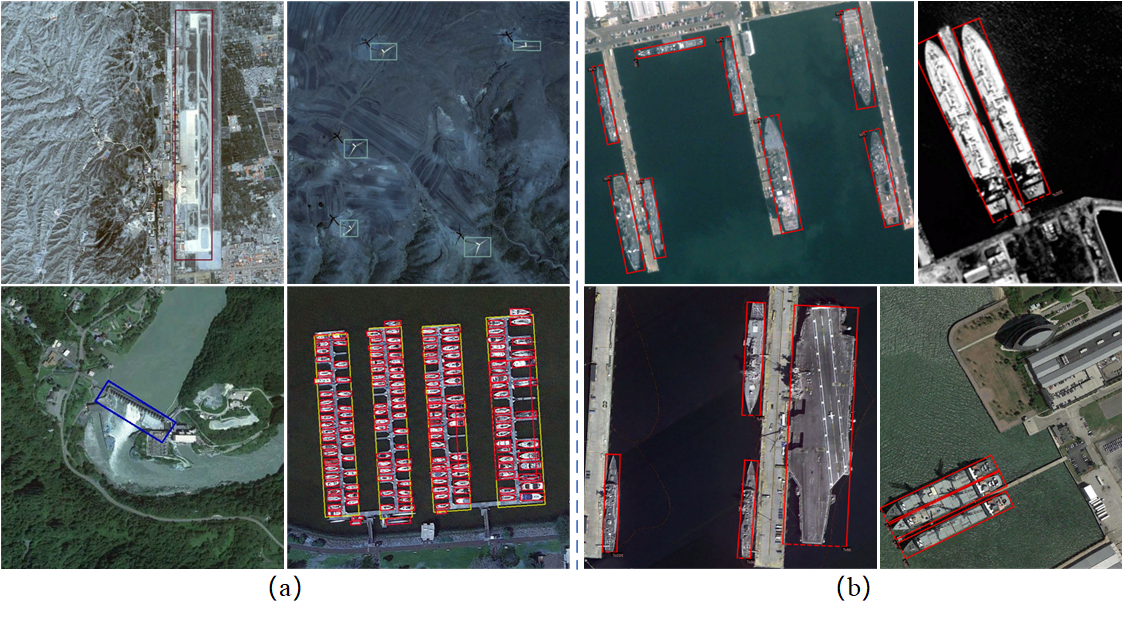}
    \caption{Visualization of detection results on  (a) DIOR-R dataset and (b) HRSC2016 dataset with our method. Best viewed in colour.}
    \label{fig:visualother}
\end{figure}
\subsubsection{Performance Evaluation on Other Datasets}
To further evaluate the robustness of the proposed EARL strategy, comparative experiments on two other commonly used RSIs datasets, i.e., DIOR-R and HRSC2016 datasets, are conducted, and the results are listed in Tables \ref{tab:dior-r} and \ref{tab:hrsc2016}, respectively.

As shown in Table \ref{tab:dior-r}, our EARL strategy with the simple anchor-free detector, i.e., R-FCOS-EARL, achieves outstanding performance of 63.3\% mAP on DIOR-R dataset, which has surpassed several state-of-the-art anchor-based detectors, such as S$^2$A-Net, R$^3$Det and Gliding Vertex. In addition, RTMDet-R-EARL achieves the pretty high mAP of 68.2\%, which further demonstrates the effectiveness and robustness of our EARL for oriented object detection in RSIs.

Furthermore, as shown in Table \ref{tab:hrsc2016}, when combined with the simple detector, our method (i.e., R-FCOS-EARL) achieves the promising performance of 89.0\% mAP$_{07}$ and 93.0\% mAP$_{12}$ on HRSC2016 dataset, compared with both anchor-based and anchor-free methods. The results indicate that R-FCOS-EARL is robust to detect targets with arbitrary orientations and large aspect ratios.
In addition, our R-FCOS-EARL with ResNet101 achieves 88.8\% mAP$_{07}$ and 92.1\% mAP$_{12}$, which exceeds other compared methods but has a slight performance degradation over ResNet50, since ResNet101 requires more training data to fully capture the model parameters, while HRSC2016 is a small-scale dataset. Furthermore, when combined with the advanced detector, RTMDet-R-EARL achieves the higher performance, i.e., 90.3\% mAP$_{07}$ and 98.0\% mAP$_{12}$, which further illustrates the robustness and generalizability of our EARL strategy in detecting targets with large aspect ratios. 
The visualization of detection results on DIOR-R and HRSC2016 datasets are illustrated in Fig.~\ref{fig:visualother}, which intuitively demonstrates the effectiveness and robustness of the proposed EARL. 

\begin{figure}[t]
    \centering
    \includegraphics[width=0.8\linewidth]{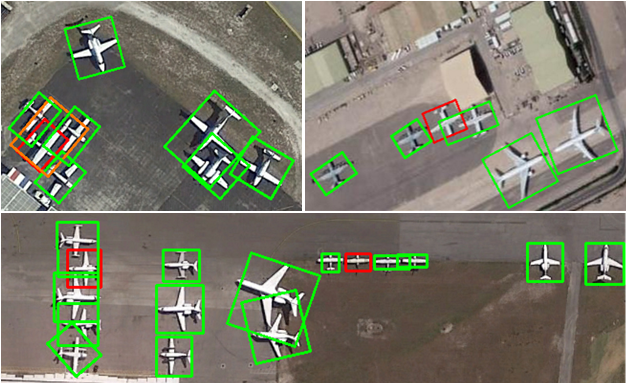}
    \caption{Illustration of failed detection results using R-FCOS-EARL. Here, the green, orange and red boxes denote true positive (TP), false positive (FP) and false negative (FN) predictions, respectively. Best viewed in colour.}
    \label{fig:failed}
\end{figure}

Though our EARL can yield better performance through sufficient and balanced sampling by considering large variations in scales and aspect ratios of targets in RSIs, it still suffers from failed detection for densely arranged targets, as illustrated in Fig. \ref{fig:failed}, where targets with dense distribution are miss detected. One possible solution to mitigate this phenomenon is to incorporate our EARL with a proprietary detector or loss function for object detection in crowd scenes, as our EARL is so simple and can be easily deployed on different models.

\section{Conclusion}
In this paper, we presented a novel and effective label assignment strategy, namely EARL, for object detection in RSIs. In which, the ADS strategy was proposed to solve the problem of insufficient and imbalanced sampling for targets with large variations in scales and aspect ratios, by selecting samples from suitable and continuous multi-level feature maps. Considering the large aspect ratios of targets, the DED strategy was designed to obtain a more accurate and reasonable sampling range while filtering out low-quality samples, which can dynamically adjust sampling range according to the aspect ratios of targets. To make the detector more focused on high-quality samples, the SDW module was developed. The experimental results demonstrate the effectiveness and robustness of our proposed EARL on three publicly available and challenging RSIs datasets, including DOTA, DIOR-R and HRSC2016. Our EARL is so simple that it has the potential to be easily combined with different orientation detectors to achieve better performance.

\bibliographystyle{IEEEtran}
\bibliography{ref}

\vfill

\end{document}